\documentclass[sigconf,authorversion]{acmart}
\usepackage{pifont}
\usepackage{bbm}
\usepackage{fancyhdr}
\usepackage{subfigure}
\usepackage{amsmath,amsfonts}
\usepackage{algorithmic}
\usepackage{algorithm}
\usepackage{textcomp}
\usepackage{stfloats}
\usepackage{verbatim}
\usepackage{multirow}
\usepackage{booktabs}
\usepackage{color}
\usepackage{enumerate}

\AtBeginDocument{%
  }

\setcopyright{acmlicensed}
\copyrightyear{2026}
\acmYear{2026}
\setcopyright{cc}
\setcctype{by}
\acmConference[KDD '26]{Proceedings of the 32nd ACM SIGKDD Conference on Knowledge Discovery and Data Mining V.1}{August 09--13, 2026}{Jeju Island, Republic of Korea}
\acmBooktitle{Proceedings of the 32nd ACM SIGKDD Conference on Knowledge Discovery and Data Mining V.1 (KDD '26), August 09--13, 2026, Jeju Island, Republic of Korea}
\acmPrice{}
\acmDOI{10.1145/3770854.3780207}
\acmISBN{979-8-4007-2258-5/2026/08}

\begin{document}

\title{Graph Defense Diffusion Model}

\author{Xin He}
\email{hex23@mails.jlu.edu.cn}
\affiliation{%
  \institution{Jilin University}
  \city{Changchun}
  \country{China}
}

\author{Wenqi Fan}
\email{wenqifan03@gmail.com}
\affiliation{%
  \institution{The Hong Kong Polytechnic University}
  \city{Hong Kong SAR}
  \country{China}
}

\author{Yili Wang}
\email{wangyili@jlu.edu.cn}
\affiliation{%
  \institution{Jilin University}
  \city{Changchun}
  \country{China}
}

\author{Chengyi Liu}
\email{chengyi.liu@connect.polyu.hk}
\affiliation{%
  \institution{The Hong Kong Polytechnic University}
  \city{Hong Kong SAR}
  \country{China}
}

\author{Rui Miao}
\email{miaorui24@mails.jlu.edu.cn}
\affiliation{%
  \institution{Jilin University}
  \city{Changchun}
  \country{China}
}

\author{Xin Juan}
\email{junxin22@mails.jlu.edu.cn}
\affiliation{%
  \institution{Jilin University}
  \city{Changchun}
  \country{China}
}

\author{Xin Wang}
\authornote{Corresponding author.}
\email{xinwang@jlu.edu.cn}
\affiliation{%
  \institution{Jilin University}
  \city{Changchun}
  \country{China}
}

\renewcommand{\shortauthors}{Xin He et al.}
\begin{abstract}
  Graph Neural Networks (GNNs) are highly vulnerable to adversarial attacks, which can greatly degrade their performance. 
  Existing graph purification methods attempt to address this issue by filtering attacked graphs. However, they struggle to defend effectively against multiple types of adversarial attacks (e.g., targeted attacks and non-targeted attacks) simultaneously due to limited flexibility. Additionally, these methods lack comprehensive modeling of graph data, relying heavily on heuristic prior knowledge.
  To overcome these challenges, we introduce the 	\textbf{G}raph \textbf{D}efense \textbf{D}iffusion \textbf{M}odel (GDDM), a flexible purification method that leverages the denoising and modeling capabilities of diffusion models. 
  The iterative nature of diffusion models aligns well with the stepwise process of adversarial attacks, making them particularly suitable for defense. 
  By iteratively adding and removing noises (edges), GDDM effectively purifies attacked graphs, restoring their original structures and features.
  Our GDDM consists of two key components: (1) Graph Structure-Driven Refiner, which preserves the basic fidelity of the graph during the denoising process, and ensures that the generated graph remains consistent with the original scope; and (2) Node Feature-Constrained Regularizer, which removes residual impurities from the denoised graph, further enhancing the purification effect. 
  By designing tailored denoising strategies to handle different types of adversarial attacks, we improve the GDDM's adaptability to various attack scenarios. 
  Furthermore, GDDM demonstrates strong scalability, leveraging its structural properties to seamlessly transfer across similar datasets without retraining.
  Extensive experiments on three real-world datasets demonstrate that GDDM outperforms state-of-the-art methods in defending against various adversarial attacks, showcasing its robustness and effectiveness.
% The code is in \href{https://github.com/hexin5515/GDDM}{https://github.com/hexin5515/GDDM}.
\end{abstract}

\begin{CCSXML}
<ccs2012>
   <concept>
       <concept_id>10002950.10003624.10003633.10010917</concept_id>
       <concept_desc>Mathematics of computing~Graph algorithms</concept_desc>
       <concept_significance>500</concept_significance>
       </concept>
   <concept>
       <concept_id>10002951.10003227.10003351.10003218</concept_id>
       <concept_desc>Information systems~Data cleaning</concept_desc>
       <concept_significance>500</concept_significance>
       </concept>
 </ccs2012>
\end{CCSXML}

\ccsdesc[500]{Mathematics of computing~Graph algorithms}
\ccsdesc[500]{Information systems~Data cleaning}
% <concept_id>10002950.10003624.10003633.10010917</concept_id>
% <concept_desc>Mathematics of computing~Graph algorithms</concept_desc>
% <concept_id>10002951.10003227.10003351.10003218</concept_id>
% <concept_desc>Information systems~Data cleaning</concept_desc>

% \ccsdesc[500]{Mathematics of computing~Graph algorithms}
% \ccsdesc[500]{Information systems~Data cleaning}

\keywords{Graph Neural Networks, Adversarial
Attack, Diffusion Model}


\maketitle
\newcommand\kddavailabilityurl{https://doi.org/10.1145/3770854.3780207}
\ifdefempty{\kddavailabilityurl}{}{
\begingroup\small\noindent\raggedright\textbf{Resource Availability:}\\
% please change the following context to include multiple artifacts if necessary, including data, models, code, etc.
The source code of this paper has been made publicly available at \url{https://doi.org/10.5281/zenodo.18028436}.
\endgroup
}

\section{INTRODUCTION}

Graph data, characterized by its ability to capture complex relationships within real-world systems, has become increasingly prevalent across diverse domains, such as recommendation systems~\cite{fan2022graph}, biological networks~\cite{wang2025adagcl+}, and social sciences~\cite{fan2019graph, derr2020epidemic}. 
Building on the potential of graph data, GNNs have emerged as a particularly effective approach to representation learning, leveraging message-passing mechanisms to aggregate both structural and feature information. This capability has enabled GNNs to achieve considerable success in downstream tasks such as node classification, link prediction, and graph classification~\cite{miao2024rethinking,shen2024optimizing}. However, it also makes them vulnerable to adversarial attacks that exploit their dependence on interconnected features and structures~\cite{jin2023empowering}.
% The complexity and richness of graph structures provide a unique opportunity to extract meaningful insights from interconnected entities. However, the same characteristics that make graph data valuable also make it vulnerable to adversarial attacks.
% Minor changes to graph structures or node features can significantly impact model outcomes, challenging the reliability of graph-based learning systems.

% Subtle modifications to graph structures or node features can drastically alter model outcomes, posing significant challenges in maintaining the reliability of graphh-based learning systems.

Adversarial attacks introduce imperceptible modifications to graph structures or node features, leading to significant degradation in model performance. Such attacks can be categorized into targeted attacks~\cite{wu2019adversarial}, which focus on misclassifying specific nodes, and non-targeted attacks~\cite{liu2023towards}, which aim to disrupt the overall graph integrity. 
The ability of these attacks to compromise model reliability poses serious challenges, especially in critical domains such as fraud detection~\cite{pourhabibi2020fraud}, healthcare~\cite{taylor2014understanding}, and cyber security~\cite{lallie2020review}, where the consequences of incorrect predictions can be severe.

\begin{figure}[!t]
  \includegraphics[width=1.0\linewidth]{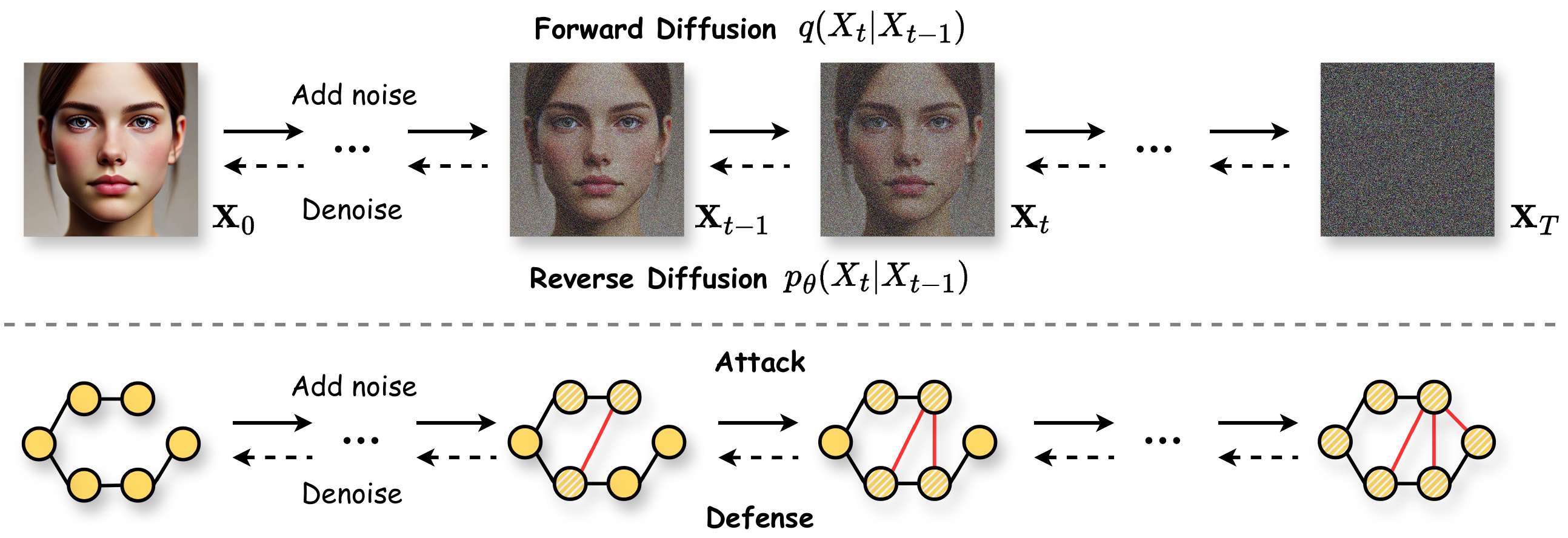}
  % \vspace{-6mm}
  \caption{The connection between graph adversarial attack, defense and diffusion model.}
\Description{}\label{fig:image_diffusion_and_graph_attack_defense}
% \vspace{-6mm}
\end{figure}

To mitigate the impact of adversarial attacks on GNNs, effective defense mechanisms are required to strengthen their robustness. Graph purification methods have shown promise by reconstructing attacked graphs without modifying the underlying GNN model. These methods are efficient and avoid the complexity of adversarial training routines~\cite{jin2020graph}. However, current purification methods lack adaptability to diverse attack strategies~\cite{li2023guard}, and depend heavily on heuristic-based prior knowledge rather than comprehensive modeling of the graph distribution~\cite{wu2019adversarial,in2024self}.
Diffusion models~\cite{peebles2023scalable,liu2023generative} represent a promising strategy for addressing challenges associated with adversarial attacks on graphs, drawing on \textbf{their inherent consistency with the process of adversarial attack and defense}. 
As shown in Fig.~\ref{fig:image_diffusion_and_graph_attack_defense}, the forward diffusion process perturbs the data with random noise, stimulating the adversarial attacks to disrupt the graphs. Correspondingly, the reverse process reconstructs samples from noise, whose objective aligns well with the defense methods.
This natural compatibility motivates us to \textbf{leverage the powerful denoising capabilities of diffusion models to defend against graph adversarial attacks uniformly.}
% Diffusion models present a viable approach for addressing certain challenges posed by adversarial attacks on graphs. These models, which have shown success in tasks like image generation and denoising, operate through a stepwise process of adding and removing noise to progressively restore corrupted data~\cite{peebles2023scalable,gao2023implicit,nie2022diffusion}. 
% \cy{This iterative nature aligns with the adversarial attack and defense mechanisms, where perturbations are introduced incrementally and need to be reversed in a similar manner.}
% As shown in Fig.~\ref{fig:image_diffusion_and_graph_attack_defense} the forward diffusion process introduces noise, similar to how adversarial attacks disrupt graphs. The reverse process then removes this noise to recover the original data, aligning well with graph defense methods. 

Although diffusion models hold potential for defending against graph adversarial attacks, their application remains unexplored due to three key challenges:
\#1 \textbf{Graph fidelity}: purifying high-fidelity graphs directly is difficult due to the complexity of graph topology and the inherent minor noise in the original graph ~\cite{sun2022adversarial,jin2021adversarial}.
\#2 \textbf{Localized denoising}: current diffusion models perform global denoising, making it hard to focus on specific nodes or edges affected by targeted attacks.
\#3 \textbf{Scalability}: current graph diffusion models~\cite{wang2024mixed,kong2023autoregressive} suffer from high computational complexity, resulting in an unacceptable training cost on large graphs.

% To address the question of developing a versatile and effective framework that adapts to different adversarial scenarios and accurately reconstructs clean graphs, 
% To address the aforementioned issues,
% we propose the \textbf{G}raph \textbf{D}iffusion \textbf{D}efense \textbf{M}odel (GDDM).
% GDDM enhances the diffusion model's \textbf{inference phase} to solve these two key challenges.
% % : maintaining structural consistency and achieving localized denoising.
% Specifically, GDDM includes two components tailored for the first challenge:
% \textbf{Graph Structure-Driven Refiner~(GSDR)} prevents the introduction of incorrect edges or the loss of critical connections during adversarial perturbations, ensuring that the structural integrity of the graph is preserved throughout the generation process. \textbf{Node Feature-Constrained Regularizer~(NFCR)} removes residual impurities from the denoised graph, further enhancing the effectiveness of graph denoising.
% % focuses on localized denoising by removing adversarial perturbations at the node feature level, enabling precise correction without affecting the overall graph structure.
% Meanwhile, GDDM provides \textbf{Tailored Attack-specific Denoising Strategies~(TADS)} for the second challenge: retaining edges unrelated to the target nodes as the initial input for the diffusion model, thereby achieving localized denoising.

In this paper, we propose the \textbf{G}raph \textbf{D}efense \textbf{D}iffusion \textbf{M}odel (GDDM) to leverage the denoising power of graph diffusion models in defending against diverse adversarial attacks on graphs, aiming to address the three aforementioned challenges. Supervised by different levels of graph labels (e.g., adjacency matrix and node feature), GDDM implements targeted interventions on the current formation state to maximize its fidelity to the ground truth: During the denoising process, \textbf{Graph Structure-Driven Refiner (GSDR)} iteratively removes incorrect edges while preserving valid connections, ensuring basic fidelity. At the end of the denoising process, \textbf{Node Feature-Constrained Regularizer (NFCR)} uses node features to eliminate residual impurities, making precise corrections without altering the overall graph structure.
% \textbf{Graph Structure-Driven Refiner~(GSDR)} filters out the incorrect edges step-by-step while preserving plausible connections, serving as the guarantee of the basic fidelity. At the end of the denoising process, \textbf{Node Feature-Constrained Regularizer~(NFCR)} utilizes node features to remove residual impurities unintentionally introduced during denoising, enabling precise corrections without compromising the overall graph structure. 
The coupling of these two components provides reliable quality control throughout the entire denoising process, collaboratively ensuring graph fidelity (Challenge \#1). 
We also design \textbf{Tailored Attack-specific Denoising Strategy~(TADS)} to focus the model on denoising edges connected to target nodes, while retaining non-target edges as the foundation of the process, achieving localized denoising (Challenge \#2).
% to retain non-target edges as the foundation of the denoising process, achieving localized denoising (Challenge 2).
% Furthermore, GDDM can effortlessly transfer from smaller datasets to larger ones for denoising without retraining, as it does not depend on the specific feature dimensionality of nodes in the graph dataset (Challenge 3).
Furthermore, GDDM uses node degree information as input to model the graph generation process, and the degree distribution of nodes often exhibits similarity across different graph datasets. Therefore, GDDM can effortlessly transfer from smaller datasets to larger ones for denoising without retraining (Challenge \#3).

% In the \textbf{training phase}, we simplify the GDDM by focusing on the critical elements necessary for defending against adversarial attacks, which reduces the model's complexity while retaining its effectiveness. Instead of learning the entire process of reconstructing the graph, we take advantage of sampling techniques to calculate intermediate steps.

The contributions of this work are summarized as follows:
\begin{itemize}
\item 
% \textbf{Novel Graph Defense Framework:} We propose the Graph Defense Diffusion Model (GDDM), a diffusion-based approach for graph adversarial attacks while ensuring graph fidelity, localized denoising and scalability.
We propose the Graph Defense Diffusion Model (GDDM), a novel graph defense framework leveraging the natural alignment between the noise adding or denoising process of diffusion models and adversarial attack or defense to effectively defend against graph adversarial attacks.

% for effectively defending against graph adversarial attacks.

\item 
% \textbf{Graph Fidelity Enhancement Components:} We introduce the Graph Structure-Driven Refiner (GSDR) and the Node Feature-Constrained Regularizer (NFCR) for high-fidelity denoised graphs across different graph label levels.
We introduce two key components (GSDR, NFCR) for high-fidelity denoised graphs and design tailored denoising strategies (TADS) to achieve localized denoising for targeted attacks. 
Besides, the similarity between graph datasets of the same type and the unique structure of the model contribute to GDDM's scalability across different datasets.                                          
% Besides, the similarity between graph datasets of the same type and the unique structure of the model enables GDDM to scale effectively across different datasets.

% \item \textbf{Tailored Denoising Strategies}: We design Tailored Attack-Specific Denoising Strategies (TADS) to remove the edges related to the target nodes from the initial adjacency matrix in targeted attack scenarios, achieving localized denoising.

% \item \textbf{Strong Scalability:} GDDM uses node degree information as input and exploits similar degree distributions across graphs, enabling seamless transfer from small to large datasets without retraining.

% \item \textbf{Simplified Training Phase:} Although the training phase is simplified by sampling techniques, our main contributions lie in the enhanced inference phase, enabling GDDM to perform robustly against various adversarial scenarios.

\item Extensive experiments demonstrate that GDDM provides superior performance in defending against adversarial attacks, outperforming current state-of-the-art methods.
\end{itemize}
\section{Related Work}

\subsection{Adversarial Attack on Graphs} Graph Neural Networks (GNNs) have been highly successful in graph data mining, but they are vulnerable to adversarial perturbations due to their dependency on the graph structure and message-passing mechanism~\cite{liu2023towards}. 
One type of adversarial attack targets specific nodes in the graph, known as targeted attacks. 
% IG-FGSM~\cite{wu2019adversarial} is an integrated gradients-based method that outperforms existing iterative and gradient-based techniques. 
Nettack~\cite{zugner2018adversarial} uses a greedy approach to perturb targeted nodes without altering node degree or feature co-occurrence.
Another type of attack targets the entire graph, known as non-targeted attacks. Mettack~\cite{DBLP:conf/iclr/ZugnerG19} is a classic method using meta-learning for global graph attacks, but it is inefficient due to its focus on low-confidence nodes. GraD~\cite{liu2023towards} improves the attack by focusing on medium-confidence nodes, enhancing efficiency. Basides, TDGIA~\cite{zou2021tdgia} proposes a gradient-based node injection strategy that has shown stronger attack capability than traditional structure perturbation attacks. To support standardized evaluation of such attacks, GRB~\cite{zheng2021graph} introduces a unified benchmark covering multiple adversarial settings.
% \vspace{-2mm}
\subsection{Graph Defense Methods} To counter adversarial attacks, various defense strategies have been proposed~\cite{zhu2019robust,yu2021graph,jin2021node,zhang2020gnnguard}. 
One type of effective defense method is graph purification.
Jaccard~\cite{wu2019adversarial} defends against adversarial attacks by pruning edges based on the feature similarity between the two connected nodes.
SVD~\cite{entezari2020all} defends against adversarial attacks by removing the high-rank perturbations introduced by the attack methods.
Guard~\cite{li2023guard} preemptively removes all potential noisy edges by node degree information to protect target nodes.
Another type of method is GNN-improved, which improves the robustness of GNNs by enhancing the GNN models themselves.
ProGNN~\cite{jin2020graph} and STABLE~\cite{li2022reliable} mitigate the negative effects of adversarial attacks on GNNs by graph structure learning to filter out the noisy edges.
% Elastic~\cite{liu2021elastic} allows the model to tolerate more noisy edges in the graph, making it more robust against adversarial attacks on graph data.
DRAGON~\cite{yuan2024mitigating} proposes a scalable framework to address robustness degradation when scaling graph learning to large datasets. And GAME~\cite{zhang2022chasing} focuses on model design and optimization strategies to improve the overall robustness of graph representations.
Although existing defense methods achieve significant success, graph purification methods are unable to defend against multiple types of adversarial attacks.
% One effective defense method is graph purification. Jaccard~\cite{wu2019adversarial} prunes edges based on feature similarity, while SVD~\cite{entezari2020all} removes high-rank perturbations. Guard~\cite{li2023guard} removes noisy edges using node degree information to protect target nodes.
% Another approach focuses on improving GNNs' robustness. ProGNN~\cite{jin2020graph} and STABLE~\cite{li2022reliable} use graph structure learning to filter out noisy edges, and Elastic~\cite{liu2021elastic} allows tolerance for noisy edges, enhancing robustness. Despite their success, existing graph purification methods cannot defend against multiple types of adversarial attacks.
% Graph purification methods like Jaccard~\cite{wu2019adversarial} and SVD~\cite{entezari2020all} focus on removing adversarial perturbations by pruning edges based on node similarity or rank. GNN-improvement approaches like ProGNN~\cite{jin2020graph} and STABLE~\cite{li2022reliable} enhance model robustness by learning better graph structures. Despite their success, these methods often struggle to defend against multiple types of attacks simultaneously.
% \vspace{-2mm}
\subsection{Diffusion Models for Graph Data} Diffusion models~\cite{xia2023diffir,liu2025score} have shown success in generating molecular and material graphs.
GDSS~\cite{jo2022score} models the joint distribution of nodes and edges using stochastic differential equations to generate desired molecular graphs.
DiGress~\cite{vignac2022digress} models a discrete diffusion process, performing graph denoising by changing types of edges at each recovery step.
EDGE~\cite{chen2023efficient} models the change of each edge in the diffusion process as a Bernoulli distribution and introduces a guidance mechanism during recovery, significantly improving training efficiency and enabling the possibility of training diffusion models on large graphs.
Although diffusion models excel at denoising, they have yet to be applied to adversarial graph defense. We observe a natural alignment between their processes and the dynamics of adversarial attacks and defenses, proposing a graph diffusion model with different denoising strategies guided by two fidelity enhancement components to defend against different types of graph adversarial attacks.
% \vspace{-2mm}
\section{PRELIMINARY}
\subsection{Problem Statement}
Let $\mathcal{G} = (\mathcal{V}, \mathcal{E}, \mathbf{A}, \mathbf{X})$ be an undirected graph, where $\mathcal{V} = \{v_1, \dots, v_N\}$ is the set of $N$ nodes and $\mathcal{E}$ is the set of edges. 
The adjacency matrix $\mathbf{A} \in \mathbb{R}^{N \times N}$ represents the connectivity between nodes, where $\mathbf{A}_{ij} = 1$ if there is an edge between node $v_i$ and node $v_j$, and $\mathbf{A}_{ij} = 0$ otherwise.
The node feature matrix $\mathbf{X} \in \mathbb{R}^{N \times d}$ contains a $d$-dimensional feature vector for each node.
Thus, a graph can also be compactly represented as $\mathcal{G} = (\mathbf{A}, \mathbf{X})$, where $\mathbf{A}$ encapsulates structural information and $\mathbf{X}$ represents node features.
In the context of node classification tasks, adversarial attacks introduce perturbations primarily into the adjacency matrix $\mathbf{A}$ and sometimes into the feature matrix $\mathbf{X}$, which leads to a degradation in the performance of GNN classifiers. 
Let $\mathcal{G}' = (\mathbf{A}', \mathbf{X}')$ be the perturbed graph, where $\mathbf{A}'$ and $\mathbf{X}'$ denote the adversarially perturbed adjacency and feature matrices, respectively.
In this work, we specifically consider the scenario where the feature matrix $\mathbf{X}$ remains unchanged, i.e., $\mathcal{G}' = (\mathbf{A}', \mathbf{X})$, and only the adjacency matrix $\mathbf{A}$ is compromised by adversarial edges. The goal is to employ a graph diffusion model to remove as many adversarial edges as possible from the perturbed graph $\mathcal{G}'$, effectively denoising the attacked graph while preserving its essential structure, thus improving the accuracy of downstream GNN classifiers in the node classification task.

% \vspace{-4mm}
\subsection{Discrete Graph Diffusion and Recovery}
The Discrete Graph Diffusion Model represents the noise injection process using a Binomial distribution~\cite{sun2023difusco}, based on the assumption that each edge modification during noise injection is independent. In this setup, the adjacency matrix $\mathbf{A}\in\mathbb{R}^{N\times{N}}$ of the target graph $\mathcal{G}$ is a binary matrix with zeros on the diagonal.
Let $\mathbf{A}^{0}$ denote the original adjacency matrix of the target graph.
We then define the diffusion process as $q(\mathbf{A}^{t}|\mathbf{A}^{t-1})$ with $t$ denoting the timestep.
In the forward diffusion process, a Bernoulli distribution $\mathcal{B}(x;\mu)$ over the binary variable $x$ with probability $\mu$ is used to represent the probability of each edge being resampled at the current timestep $t$:
\begin{align}
    q(\mathbf{A}^{t}|\mathbf{A}^{t-1})
    &=\begin{matrix} \prod_{i,j} \mathcal{B}(\mathbf{A}_{i,j}^{t};(1-\beta_{t})\mathbf{A}_{i,j}^{t-1}+\beta_{t}p)
    \end{matrix}\nonumber
    \\
    &=\mathcal{B}(\mathbf{A}^{t};(1-\beta_{t})\mathbf{A}^{t-1}+\beta_{t}p),
\end{align}
where $\beta_{t}$ is the diffusion rate controlling the probability of resampling the element $\mathbf{A}_{i,j}^{t}$ from the Bernoulli distribution with probability $p$, rather than keeping the original $\mathbf{A}_{i,j}^{t}$.
Therefore, this specific diffusion process allows direct sampling of $\mathbf{A}^{t}$ from $\mathbf{A}^{0}$:
\begin{align}\label{eq:general_foward_diffusion}     q(\mathbf{A}^{t}|\mathbf{A}^{0})=\mathcal{B}(\mathbf{A}^{t};\bar{\alpha}_{t}\mathbf{A}^{0}+(1-\bar{\alpha}_{t})p),
\end{align}
where $\alpha_{t}=1-\beta_{t}$ and $\bar{\alpha}_{t}=\prod_{i=1}^{t}\alpha_{i}$.
As $t$ approaches infinity, $\bar{\alpha}_{t}$  converges to $0$, meaning that the final adjacency matrix $\mathbf{A}^{T}$ becomes fully independent of the original
$\mathbf{A}^{0}$. The value of $p$ controls the probability of forming edges in $\mathbf{A}^{T}$, where a higher value of $p$ implies a denser graph.

During the denoising process, the goal is to reverse the diffusion process by reconstructing $\mathbf{A}^{t-1}$ from $\mathbf{A}^{t}$, using a trained model to approximate the posterior distribution:
\begin{align}
        q(\mathbf{A}^{t-1}|\mathbf{A}^{t},\mathbf{A}^{0})=\frac{q(\mathbf{A}^{t}|\mathbf{A}^{t-1})q(\mathbf{A}^{t-1}|\mathbf{A}^{0})}{q(\mathbf{A}^{t}|\mathbf{A}^{0})}.
        \label{eq:reverseDiff}
\end{align}

Since all terms are known, the model can directly compute Eq.~\ref{eq:reverseDiff} for recovering the original graph structure during training.

\section{Graph Defense Diffusion Model}
\begin{figure*}[ht!]
  \includegraphics[width=\textwidth]{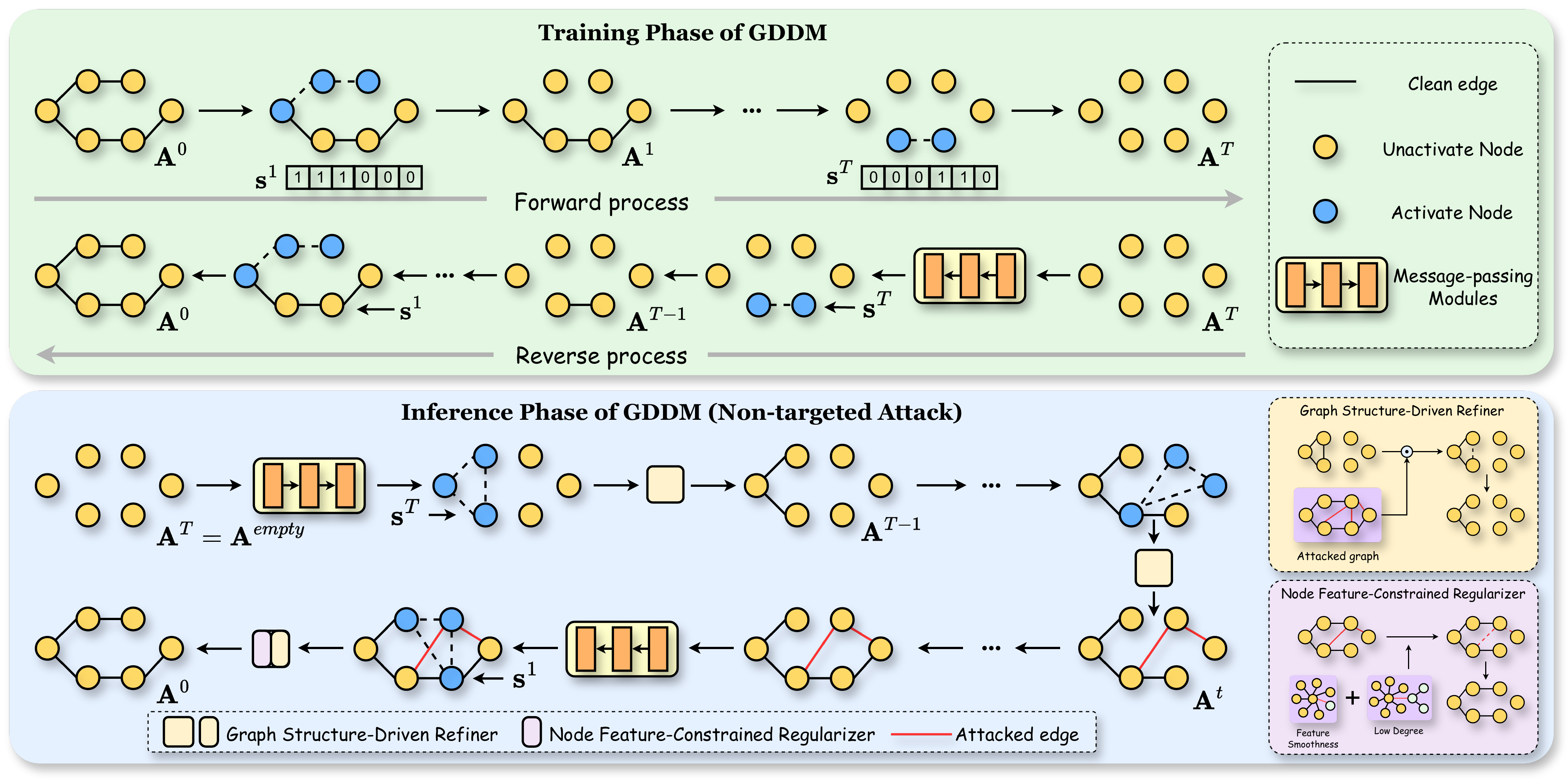}
  % \vspace{-8mm}
  \caption{The Framework of GDDM. }
  \Description{}
  \label{fig:framework}
  % \vspace{-6mm}
\end{figure*}

The architecture of the Graph Defense Diffusion Model (GDDM) is illustrated in Fig.~\ref{fig:framework}, comprising two key components: the training phase and the inference phase.
In the \textbf{training phase}, GDDM is trained on clean graph data leveraging a simplified discrete graph diffusion model. We reduce the model's complexity by employing sampling techniques for intermediate step calculations, enhancing training efficiency while maintaining the ability to reconstruct clean graph structures.
During the \textbf{inference phase}, we introduce the graph structure-driven refiner and the node feature-constrained regularizer. They ensure high-fidelity denoised graph
while enhancing GDDM's ability to defend against adversarial attacks. 
We incorporate tailored attack-specific denoising strategies to effectively address different types of adversarial attacks.
% allowing GDDM to flexibly adapt its defense mechanism based on the nature of the attack.
Additionally, GDDM leverages the similarity in degree distributions across datasets to enable transfer between similar datasets due to its unique network structure, achieving strong scalability.
In the following sections, we will provide detailed descriptions of these components.

% \vspace{-2mm}
\subsection{The Training Phase}
\subsubsection{Forward and Recovery}
Graph diffusion models inherently involve high complexity in both time and space, posing significant challenges for efficient computation.
To address this challenge while preserving the model's ability to accurately reconstruct graph structures, we simplify the model by introducing a state vector $\mathbf{s}^{t}$~\cite{chen2023efficient}, which tracks the nodes with degree changes at each timestep $t$ of the discrete graph diffusion process.
This simplification reduces computational overhead while maintaining the model's ability to restructure graphs effectively. Specifically, the state vector $\mathbf{s}^{t}$ indicates whether the degree of node $i$ changes between timestep
$t-1$ and $t$. If the degree changes, $\mathbf{s}^{t}_{i}$ is set to $1$, otherwise $0$.
The \textbf{forward diffusion process} of GDDM can be formalized as: 
\begin{align}
q(\mathbf{A}^{t},\mathbf{s}^{t}|\mathbf{A}^{t-1})=q(\mathbf{A}^{t}|\mathbf{A}^{t-1})=\mathcal{B}(\mathbf{A}^{t};(1-\beta_{t})\mathbf{A}^{t-1}+\beta_{t}p).
\end{align}
where $\beta_{t}$ controls the probability of resampling edges. 
Notably, the introduction of  $\mathbf{s}^{t}$ does not modify the core diffusion process but provides a mechanism to track which nodes' degrees are altered.
Furthermore, setting $p = 0$ transforms the forward diffusion into an edge removal process.

The \textbf{recovery process} aims to refine the perturbed graph by progressively recovering its original structure.
During the recovery process, we train a message-passing neural network $p_{\theta}(\mathbf{A}^{t-1}|\mathbf{A}^{t})$ to approximate the posterior of the forward process:
\begin{align} \label{eq:given_a0_at_st_to_at-1}
    q(\mathbf{A}^{t-1}|\mathbf{A}^{t},\mathbf{s}^{t},\mathbf{A}^{0})=\frac{q(\mathbf{A}^{t}|\mathbf{A}^{t-1})q(\mathbf{s}^{t}|\mathbf{A}^{t},\mathbf{A}^{t-1})q(\mathbf{A}^{t-1}|\mathbf{A}^{0})}{q(\mathbf{A}^{t}|\mathbf{A}^{0})q(\mathbf{s}^{t}|\mathbf{A}^{t},\mathbf{A}^{0})}.
\end{align}
Further derivation details are provided in Appendix~\ref{appendix:derivation_process_of_posterior}. In each recovery step, the network computes the state vector $\mathbf{s}^{t}$, which identifies the nodes for which edges should be generated.
% determines which nodes should have edges generated. 
The recovery process is decomposed as follows: 
% In each recovery step, $p_{\theta}(\mathbf{A}^{t-1}|\mathbf{A}^{t})$ first computes the state vector $\mathbf{s}^{t}$ to select which nodes should have edges generated, instead of generating edges indiscriminately. Then, it generates edges between the selected nodes.
% Therefore, $p_{\theta}(\mathbf{A}^{t-1}|\mathbf{A}^{t})$ can be decomposed as follows:
\begin{align}\label{eq:get_At-1_given_At}
    p_{\theta}(\mathbf{A}^{t-1}|\mathbf{A}^{t})=p_{\theta}(\mathbf{A}^{t-1}|\mathbf{s}^{t},\mathbf{A}^{t})p_{\theta}(\mathbf{s}^{t}|\mathbf{A}^{t}).
\end{align}

Although this decomposition introduces new optimization objectives, previous work~\cite{wu2023edge++} has shown a strong correlation between the state vector $\mathbf{s}^{t}$ and the degree vectors $\mathbf{d}^{0}$ and $\mathbf{d}^{t}$. The state vector can be predicted as:
% Although this decomposition method can improve the training efficiency of the model, it may also impact the quality of the final generated graph due to the introduction of new optimization objectives.
% Fortunately, previous work~\cite{wu2023edge++} found that there is a strong correlation between the state vector $\mathbf{s}^{t}$ and the degree vectors $\mathbf{d}^{0}$ and $\mathbf{d}^{t}$.
% The posterior can be formulated as follows:
\begin{align}\label{eq:pred_st_given_d0_and_dt}
q(\mathbf{s}^{t}|\mathbf{d}^{t},\mathbf{d}^{0})&=\prod_{i=1}^{N}q(\mathbf{s}^{t}_{i}|\mathbf{d}^{t}_{i},\mathbf{d}^{0}_{i}),\\
q(\mathbf{s}^{t}_{i}|\mathbf{d}^{t}_{i},\mathbf{d}^{0}_{i})&=\mathcal{B}(\mathbf{s}^{t}_{i};1-(1-\frac{\beta_{t}\bar{\alpha}_{t-1}}{1-\bar{\alpha}_{t-1}})^{\mathbf{d}^{0}_{i}-\mathbf{d}^{t}_{i}}),\nonumber
\end{align}
where $N$ represents the total number of nodes in the graph, $\alpha_{t}=1-\beta_{t}$ and $\bar{\alpha}_{t}=\prod_{i=1}^{t}\alpha_{i}$.
During the training phase, the optimization is focused only on $p_{\theta}(\mathbf{A}^{t-1}|\mathbf{s}^{t},\mathbf{A}^{t})$. The state vector $\mathbf{s}^{t}$ is directly sampled from the degree vectors $\mathbf{d}^{0}$ and $\mathbf{d}^{t}$, which simplifies the training phase by eliminating the need for additional optimization of $p_{\theta}(\mathbf{s}^{t}|\mathbf{A}^{t})$, thus improving computational efficiency without compromising model performance.
The detailed derivation of Eq.~\ref{eq:pred_st_given_d0_and_dt} can be found in the Appendix~\ref{appendix:derivation_of_eq7}.

% where $\beta_{t}$ is the diffusion rate  controlling the probability of resampling.
% $\alpha_{t}=1-\beta_{t}$ and $\bar{\alpha}_{t}=\prod_{i=1}^{t}\alpha_{i}$.
% Based on the aforementioned correlation, we only need to optimize the network $p_{\theta}(\mathbf{A}^{t-1}|\mathbf{s}^{t},\mathbf{A}^{t})$ during the training phase of GDDM, as the output of the network $p_{\theta}(\mathbf{s}^{t}|\mathbf{A}^{t})$ can be directly sampled using Eq.~\ref{eq:pred_st_given_d0_and_dt} given the known degree vectors $\mathbf{d}^{0}$ and $\mathbf{d}^{t}$.

\subsubsection{Optimization Objective}
Building on the recovery process discussed earlier, we now define the optimization objective for GDDM.
To optimize the parameters $\theta$, we maximize the variational lower bound (VLB)~\cite{ho2020denoising,sohl2015deep} of $\log p(\mathbf{A}^0)$, ensuring that the model learns to accurately reconstruct the original graph from its perturbed version. The optimization objective is defined as follows:
\begin{align}\label{training_loss}
\mathcal{L} = \mathbb{E}_{q} \left[\log\frac{p(\mathbf{A}^T)}{q(\mathbf{A}^T|\mathbf{A}^0)} + \log p_{\theta}(\mathbf{A}^0|\mathbf{A}^1,\mathbf{s}^1) + \right. \notag\\
\left. \sum_{t=2}^T\log\frac{p_{\theta}(\mathbf{A}^{t-1}|\mathbf{A}^t,\mathbf{s}^t)}{q(\mathbf{A}^{t-1}|\mathbf{A}^t,\mathbf{A}^0, \mathbf{s}^t)} \right].
\end{align}
The first term does not involve learnable parameters, while Monte Carlo estimation~\cite{hammersley2013monte} is used to optimize the second and third terms. This process enables GDDM to iteratively reconstruct the graph from an empty state, capturing its structure and improving denoising performance. The overall training phase of GDDM is presented in Appendix~\ref{appendix:training_alg}.

% The first term in the loss function does not contain learnable parameters, and we use Monte Carlo estimation~\cite{hammersley2013monte} to optimize the second and third terms.
% Through this optimization, GDDM learns to reconstruct the original graph iteratively by adding edges from an empty graph. This process allows GDDM to effectively capture the graph structure and improve its denoising capability.

\subsubsection{Model Structure}
To better understand how GDDM functions, we now provide an overview of its internal structure. GDDM consists of two main components: information propagation and edge prediction, which work together to reconstruct the graph from an empty structure iteratively. 

In the information propagation process, node degree representations are used to capture features. The initial representation of each node is constructed by concatenating the node's initial $\mathbf{d}^0$ with its current degree $\mathbf{d}^t$, along with the timestep $t$ and global contextual features $\mathbf{c}$. This can be formulated as:
\begin{align}
    \mathbf{Z}^{0}_i=&{\rm concat}({\rm emb}_{d}(\mathbf{d}^t_i/\mathbf{d}_{max})\|{\rm emb}_d(\mathbf{d}^0_i/\mathbf{d}_{max})), \\
    &\mathbf{t}_{emb} = {\rm emb}_{t}(t),\mathbf{c}^0={\rm mean}(\mathbf{Z}^0),
\end{align}
% \begin{align}
%     \mathbf{t}_{emb} = {\rm emb}_{t}(t),\mathbf{c}^0={\rm mean}(\mathbf{Z}^0),
% \end{align}
where ${\rm concat}(\cdot)$ represents the concatenation operation, ${\rm emb}_{d}(\cdot)$ and ${\rm emb}_{t}(\cdot)$ represent the degree and time encoding functions, $\mathbf{c}^0$ captures the initial global context, and $\mathbf{d}_{max}$ represents the maximum node degree in the original graph. Then, these initial node representations are updated iteratively as follows:
\begin{align}
    \textbf{Z}^{l}, \textbf{c}^{l},\textbf{H}^{l}={\rm MPM}(\textbf{Z}^{l-1},\textbf{t}_{emb},\textbf{c}^{l-1},\textbf{H}^{l-1}),
\end{align}
where $\textbf{H}$ refers to the node hidden features at each layer $l$. 
The message-passing mechanism (MPM) propagates information across the graph to refine the node representations iteratively. The detail of the information propagation process is in Appendix~\ref{appendix:detail_of_MPM}.

For edge prediction, after $L$ iterations of updates, a multi-layer perceptron (MLP) is used to predict whether an edge should be generated between nodes $i$ and $j$. The prediction is based on the final representations $\mathbf{Z}_i^L$ and $\mathbf{Z}_j^L$ of the nodes: 
\begin{align}
    {b}_{i,j}^{t-1}&={\rm MLP}(\mathbf{Z}_i^L+\mathbf{Z}^{L}_j), {\rm where}\ (i,j)\in \{(i,j)|\mathbf{s}_{i,j}^t=1 \},    
\end{align}
where $b_{i,j}^{t-1}$ is the predicted two-dimensional Bernoulli parameter for generating an edge between node $i$ and node $j$. Then, whether an edge is generated between nodes $i$ and $j$ is determined by a sampling process based on ${b}_{i,j}^{t-1}$ and a Gumbel noise~\cite{jang2016categorical} $g$ with the same dimensionality as ${b}_{i,j}^{t-1}$, which can be defined as follows:
\begin{align}
    \mathbf{A}_{i,j}^{t-1}=\arg\max({b}_{i,j}^{t-1}+g).
\end{align}

\subsection{The Inference Phase}
Following the completion of the training phase, GDDM applies the learned diffusion dynamics to reconstruct the clean graph from the attacked one. During inference, the process iteratively restores the graph by sampling the state vector $\mathbf{s}^{t}$
and adjacency matrix $\mathbf{A}^{t-1}$ at each timestep, as defined by:
% The inference phase of GDDM at each timestep can be formulated as follows:
\begin{align}\label{eq:sample_st_given_dt_d0}
    \mathbf{s}^t &\thicksim  q(\mathbf{s}^{t}|\mathbf{d}^{t},\mathbf{d}^{0}), \\ 
    \mathbf{A}^{t-1} &\thicksim p_{\theta}(\mathbf{A}^{t-1}|\mathbf{s}^{t},\mathbf{A}^{t}), \label{eq:sample_At-1_given_st_At}
\end{align}
% \begin{align}\label{eq:sample_At-1_given_st_At}
%     \mathbf{A}^{t-1} \thicksim p_{\theta}(\mathbf{A}^{t-1}|\mathbf{s}^{t},\mathbf{A}^{t}),
% \end{align}
where $\mathbf{d}^0$ is the degree vector of the attacked graph, $\mathbf{d}^t$ is the degree vector at timestep $t$, and $p_{\theta}$ represents the learnable network.

Following the sampling steps in inference, the next crucial task is to ensure that the generated graph maintains a high-fidelity with the attacked graph. This is essential to ensure that while adversarial noise is mitigated, the core structure of the original graph is preserved. To achieve this, GDDM integrates two core components: Graph Structure-Driven Refiner (GSDR) and Node Feature-Constrained Regularizer (NFCR). Additionally, tailored denoising strategies are employed to handle various types of adversarial attacks, allowing GDDM to effectively target both global and localized attacks.

% When denoising a graph, modifications need to be made based on the attacked graph. 
% This requires GDDM to ensure that the generated graph retains a high-fidelity throughout the inference phase.
% In other words, the generated graph's structure should closely resemble the original graph, or even be cleaner than the original.
% Therefore, we design two components: (1) the Graph Structure-Driven Refiner (GSDR) prunes edges to keep the generated graph as a refined version of the attacked graph, guaranteeing the basic fidelity; and (2) the Node Feature-Constrained Regularizer (NFCR) eliminates minor impurities, further enhancing the fidelity of generated graph.
% Additionally, we design tailored denoising strategies for different adversarial attacks to make GDDM achieve localized denoising for targeted attacks.
% We will introduce these modules and denoising strategies in detail below.

\subsubsection{Graph Structure-Driven Refiner (GSDR)}

GSDR is designed to tackle the challenge of generating high-fidelity graphs during the denoising process. Due to the complexity of graph topology and the inherent noise in the original graph, directly generating a high-fidelity graph is difficult. GSDR addresses this by incrementally refining the graph structure at each timestep. At each step of the generation process, GSDR filters out incorrect edges, ensuring that only edges consistent with the attacked graph are retained. This process is formalized as:
\begin{align}\label{eq:filter_At}
    \mathbf{A}^{t} = \mathbf{A}' \odot \mathbf{A}^{t},
\end{align}
where $\mathbf{A}'$ represents the adjacency matrix of the attacked graph, $\mathbf{A}^{t}$ represents the adjacency matrix of the corresponding graph at timestep $t$ during the generation process. The element-wise multiplication $\odot$ preserves the intersection between $\mathbf{A}'$ and $\mathbf{A}^{t}$.
By filtering edges step-by-step, GSDR preserves plausible connections while removing adversarial noise, ensuring the generated graph remains structurally consistent with the attacked graph.

% To ensure the aforementioned consistency, in this work, we design a Graph Structure-Driven Refiner (GSDR) to guide the generation process of GDDM.
% Specifically, at each generation step, we remove edges that do not belong to the attacked graph from the generated edges, which can be formalized as follows:
% \begin{align}\label{eq:filter_At}
%     \mathbf{A}^{t} = \mathbf{A}' \odot \mathbf{A}^{t},
% \end{align}
% where $\mathbf{A}'$ represents the adjacency matrix of the attacked graph, $\mathbf{A}^{t}$ represents the adjacency matrix of the corresponding graph at timestep $t$ during the generation process.
% The element-wise multiplication $\odot$ preserves the intersection between $\mathbf{A}'$ and $\mathbf{A}^{t}$.
% This approach ensures the graph generated at each step by GDDM remains within the scope of the attacked graph.
% Based on this approach, we only need to control the scale of the generated graph to remove noisy edges from the attacked graph.

\subsubsection{Node Feature-Constrained Regularizer (NFCR)}
While GSDR effectively removes the most noisy edges, residual noise in node features can still affect the overall graph quality. NFCR refines the generated graph by incorporating node feature and degree information, which addresses residual noise that remains after structural pruning. NFCR ensures that both node features and the underlying graph structure are preserved, enhancing the fidelity of the generated graph. NFCR operates in two stages: feature smoothness guiding and node degree guiding.

% Although the powerful generative ability of diffusion models is sufficient to remove most noisy edges in the graph, the inherent noise in the original graph can still impact GDDM's denoising effectiveness. 
% To address this limitation and refine the results, we design a Node Feature-Constrained Regularizer (NFCR) to filter the edges by node feature and degree information from the generated graph to further enhance the defense effectiveness.

\noindent\textbf{Feature smoothness guiding.}
As the generated graph reaches a certain size, it becomes necessary to evaluate whether the generated edges connect nodes that should logically be connected.
This is achieved by assessing the feature smoothness between pairs of nodes connected by the generated edges. The feature smoothness is calculated as follows:
% Once the generated graph reaches a certain scale,
% we assess whether the generated edge truly connects two nodes belonging to the same category by evaluating the feature smoothness~\cite{ando2006learning} between the two nodes connected by the generated edge.
% The calculation process for feature smoothness between any two nodes in the graph is defined as follows:
\begin{align}\label{eq:feature_smoothness}
    \mathbf{FS}(i,j) = \|\mathbf{X}_i-\mathbf{X}_j\|^2_2,
\end{align}
where $\mathbf{X}$ is the feature matrix of the original graph. 
Each row vector $\mathbf{X}_i$ or $\mathbf{X}_j$
in the feature matrix $\mathbf{X}$ represents the one-hot vector of a node,
$\|\cdot\|^2$ represents the square of the second-order norm of the vector. 
$\mathbf{FS}(i,j)$ represents the feature smoothness between the two nodes, with higher values indicating lower feature similarity and a reduced likelihood of an edge existing between them. After calculating the feature smoothness for all edges in the generated graph, the top-$k$ edges with the highest smoothness values (i.e., lowest feature similarity) are removed. This step ensures that only edges connecting nodes with similar features are retained, improving the quality of the generated graph.
% $\mathbf{FS}(i,j)$ represents the feature smoothness between node $i$ and $j$.
% A higher $\mathbf{FS}(i,j)$ indicates lower feature similarity between node $i$ and node $j$, which reduces the likelihood of an edge existing between them.
% We calculate the feature smoothness for each pair of nodes connected by the edges in the generated graph.
% Then, we remove the top-$k$ edges with the highest feature smoothness values.
% \vspace{-3mm}

\begin{figure}[t]
% \vspace{-4mm}
\centering
{
{\includegraphics[width=0.49\linewidth]{{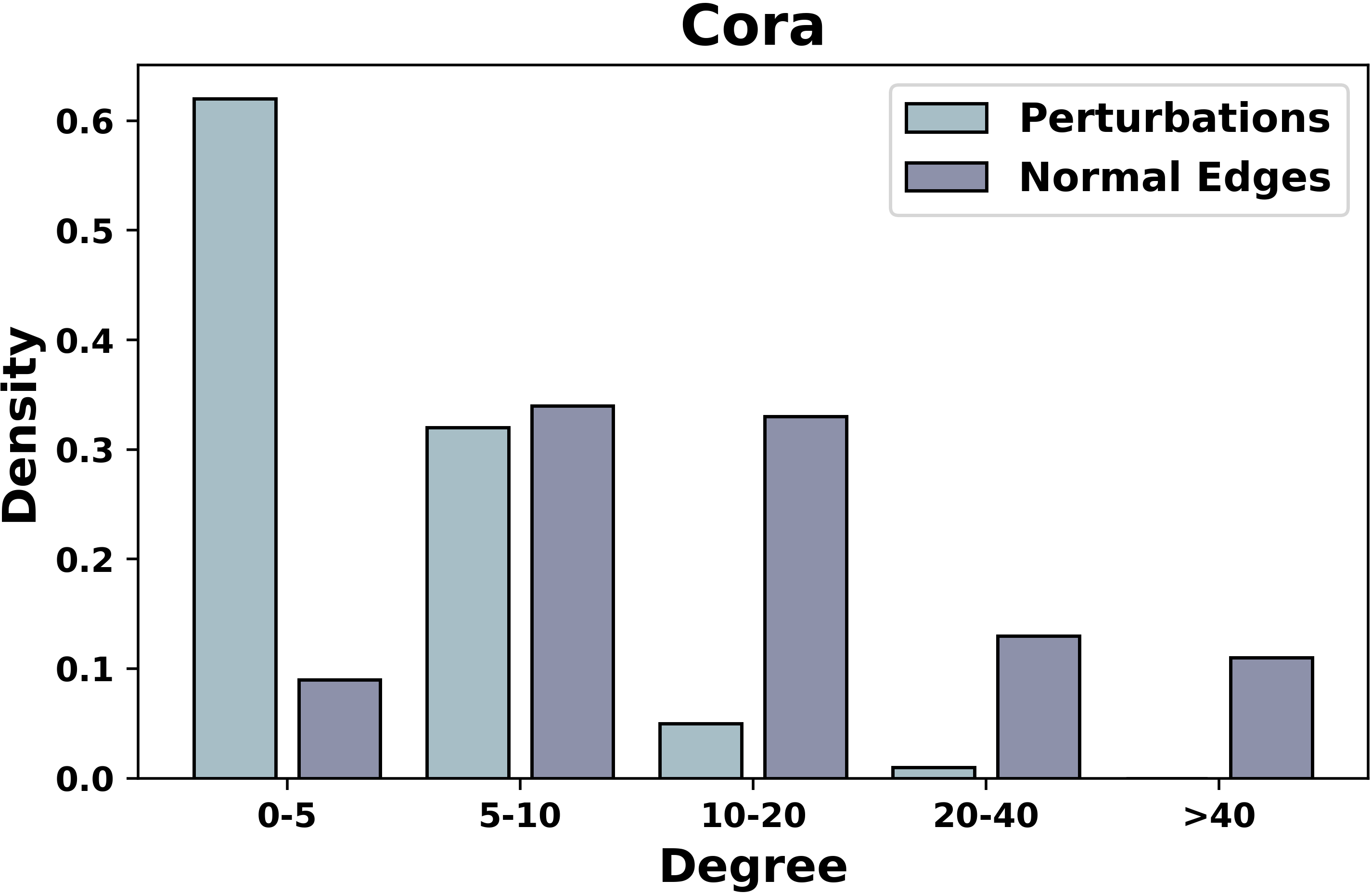}}}}
{
{\includegraphics[width=0.49\linewidth]{{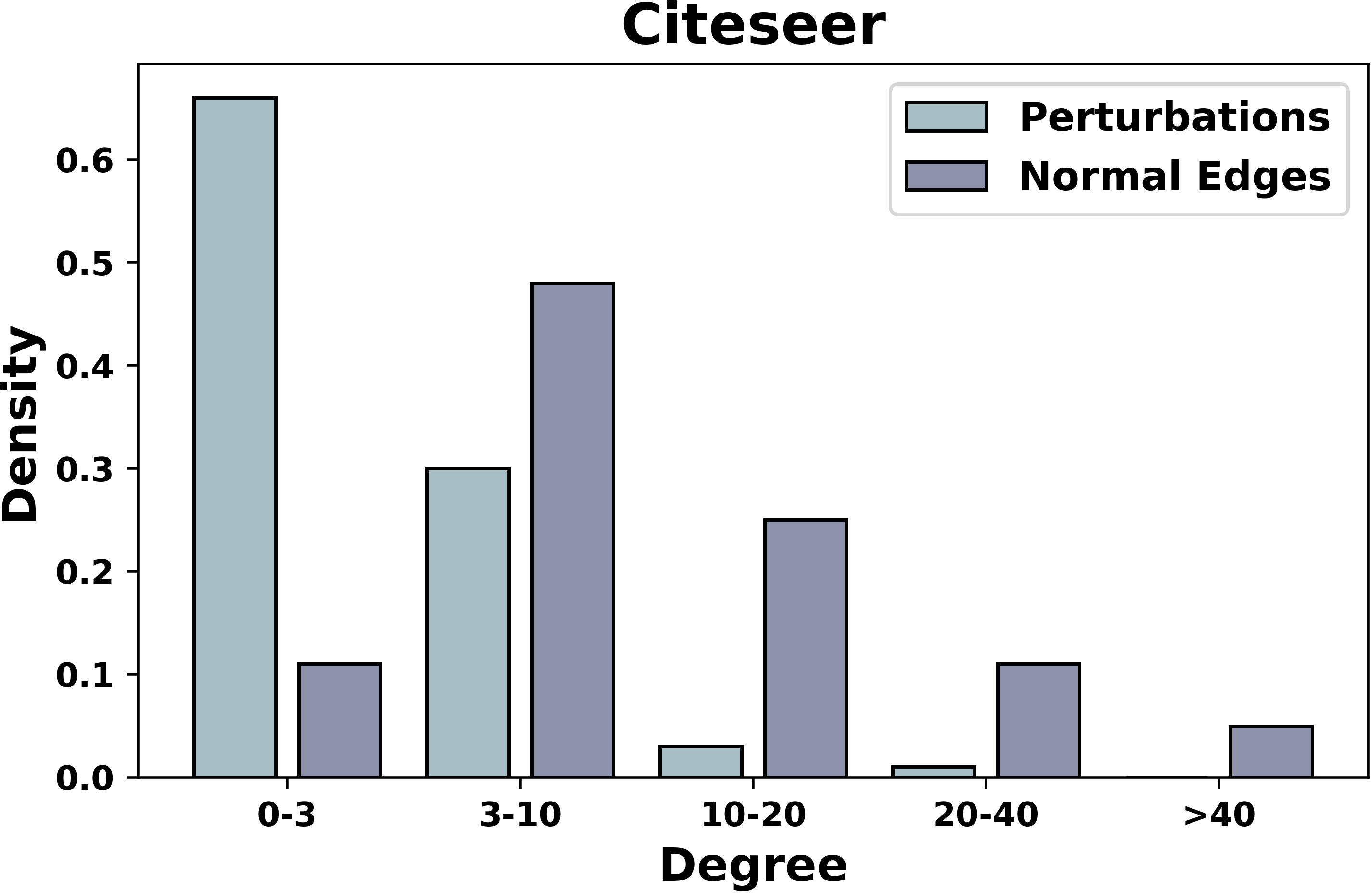}}}}%
% \vspace{-4mm}
\Description{The edge degree distribution}
\caption{The edge degree distribution of the clean graph and the distribution of adversarial edges generated by GraD on the Cora and Citeseer dataset.}\label{fig:attacked_edge_with_node_degree}
% \vspace{-6mm}
\end{figure}

\noindent\textbf{Node degree guiding.}
In addition to feature smoothness, the node degrees of the connected nodes also play a crucial role in determining whether an edge should be retained. Previous studies, such as GraD~\cite{liu2023towards}, reveal that graph attack methods tend to target low-degree nodes by introducing noisy edges that connect these vulnerable nodes. To explore this, we visualize the distribution of edges in the attacked graph, as shown in Fig.~\ref{fig:attacked_edge_with_node_degree}. 
In the visualization, green bars show the proportion of normal edges in each degree range relative to all normal edges, while orange bars indicate the proportion of attacked edges in each degree range relative to all attacked edges.
We define an edge's degree as the sum of the degrees of its two connected nodes.
% the horizontal and vertical axes correspond to the degrees of the two nodes connected by each edge in the original graph. 
As shown in the figure, attack methods frequently add edges that involve two low-degree nodes. 
This pattern indicates that low-degree nodes are more susceptible to being targeted by these attacks.

To address this vulnerability, the node degree information is integrated into the edge filtering process. Specifically, if an edge exhibits high feature smoothness (indicating a low likelihood of existence) and involves low-degree nodes, it is likely to be a noisy edge introduced by the attack. The probability of an edge being targeted by an attack is defined as: 
\begin{align}\label{eq:node_degree}
    P_{\text{attack}}(i,j) = 
    \begin{cases} 
    1 & \text{if } d(i) < \lambda \ \text{or} \ d(j) < \lambda \\
    0 & \text{else }
    \end{cases},
\end{align}
where $d(\cdot)$ represents the degree of the target node in the attacked graph. $\lambda$ represents the threshold, which varies depending on the dataset. After an edge is flagged based on node degrees, it undergoes further evaluation using feature smoothness.
% Once an edge has been flagged based on its node degrees, it is further evaluated using feature smoothness. 
If $P_{\mathbf{attack}}(i,j)=1$ and the edge rank within the top-$k$ in terms of feature smoothness, the edge is removed from the generated graph. Otherwise, the edge is retained.

By combining feature smoothness and node degree filtering, NFCR effectively removes noisy edges, especially those from low-degree nodes targeted by adversarial attacks. 
Together with GSDR, which ensures structural consistency by pruning incorrect edges, these components work to maintain the fidelity of the generated graph. While GSDR focuses on preserving the correct structure, NFCR refines node-level details, ensuring a clean and accurate graph, addressing \textbf{Challenge \#1: Graph Fidelity}.

\subsubsection{Tailored Attack-Specific Denoising Strategies (TADS)}
While GSDR and NFCR address the global fidelity of the graph, targeted attacks require more localized denoising strategies. To address \textbf{Challenge \#2: localized denoising}, GDDM adopts tailored strategies for different types of adversarial attacks. 
For \textbf{non-targeted attacks}, the adjacency matrix $\mathbf{A}^{empty}$ of an empty graph is used as the starting point, allowing GDDM to reconstruct the entire graph from scratch. This global denoising process ensures that all edges are regenerated, as shown in Fig.~\ref{fig:framework}.
For \textbf{targeted attacks}, where specific nodes are compromised, a more focused strategy is necessary. GDDM starts by removing all edges connected to the target nodes, constructing an initial adjacency matrix $\mathbf{A}^{tar}$, defined as:
\begin{align}
    \mathbf{A}^{tar} = \mathbf{A}'-\mathbf{A}'',
\end{align}
where $\mathbf{A}'$ represents the adjacency matrix of the attacked graph, $\mathbf{A}''$ represents the adjacency matrix containing only the edges connected to the target nodes. This allows GDDM to perform localized denoising, focusing on the areas most affected by the attack, as shown in Appendix~\ref{appendix:denoise_for_targeted}.

% \vspace{-2mm}
\subsection{Inherent Strong Scalability}
The impractically high training cost of graph diffusion models on large datasets is a key challenge that hinders their application in graph purification.
This limitation (\textbf{Challenge \#3: Scalability}) prevents these models from scaling to larger datasets within a short period.
Unlike existing graph diffusion models, GDDM does not rely on the specific feature dimensions of nodes in a graph dataset. 
Instead, it utilizes node degree information as the initial node features, allowing for easier transfer from smaller to larger datasets for denoising. This is facilitated by the structural similarity in degree distributions across different graph datasets (e.g., citation networks), reducing the need for retraining.
Specifically, when defending against adversarial attacks on larger datasets, the model trained on a smaller dataset can be directly used, with only the node features being reinitialized based on the degree distribution of the new graph dataset. 
We present detailed experimental results in Section~\ref{sec:rob_eva} and demonstrate that the scalability of GDDM can be successfully applied across similar datasets.
\section{Experiment}

In this section, we present our experimental setup and the empirical evaluation of GDDM's defense effectiveness and robustness against different adversarial attacks. 
% Specifically, we seek to address the following key research questions (RQs):
% \begin{itemize}
%     \item \textbf{(RQ1)} Does GDDM outperform state-of-the-art graph defense models across various adversarial attacks while maintaining strong scalability?
%     \item \textbf{(RQ2)} Does our designed denoising strategy enhance GDDM's ability to protect specific nodes under targeted attack?
%     \item \textbf{(RQ3)} Do the additional components improve the model's defense effectiveness against various attack methods?
%      \item \textbf{(RQ4)} Does GDDM integrate with different downstream GNNs and remain effective?
%     \item \textbf{(RQ5)} How does the performance of GDDM vary with different hyper-parameters?
% \end{itemize}
% \vspace{-2mm}
\subsection{Experimental Settings}
\subsubsection{Dataset.}
We conduct our experiments on three datasets 
% under \textbf{poisoning attacks}(non-targeted structure attack~\cite{liu2023towards} and targeted structure attack~\cite{zugner2018adversarial})
, consisting of Cora and Citeseer, and Pubmed. 
Notably,  we directly apply the training parameters from the smaller Cora dataset to defend against adversarial attacks on Pubmed dataset, thereby verifying the scalability of GDDM.
% When processing the Chameleon dataset, we treated it as a homogeneous dataset by considering all edges as a single category.
For each graph, we randomly split the nodes into 10\% for training, 10\% for validation, and 80\% for testing. 
The statistics for these datasets are provided in Appendix~\ref{appendix:dataset}.

\begin{table*}[ht]
\centering
% \vskip -0.3in
\caption{Node classification performance (Accuracy±Std) under targeted attack (Nettack). Results marked with * are obtained by transferring parameters from Cora dataset}
% \vspace{-4mm}
% \vskip -0.1in
\label{tab:performance_com_under_tar_att}
\scalebox{0.85}{
\begin{tabular}{c|ccccccccccc}
% \specialrule{0.08em}{0pt}{0pt}
\toprule
\multicolumn{1}{c}{\textbf{Datasets}}              & 
\multicolumn{1}{|c|}{\textbf{Ptb Num}} & \multicolumn{1}{c}{\textbf{GCN}}& \multicolumn{1}{c}{\textbf{Jaccard}}& \multicolumn{1}{c}{\textbf{SVD}}& \multicolumn{1}{c}{\textbf{ProGNN}}& \multicolumn{1}{c}{\textbf{Guard}}& \multicolumn{1}{c}{\textbf{STABLE}}& \multicolumn{1}{c}{\textbf{SG-GRS}}&\multicolumn{1}{c}{\textbf{WGTL}}&\multicolumn{1}{c}{\textbf{DiffSP}}& \multicolumn{1}{c}{\textbf{GDDM}}\\
% \hline
\midrule
\multirow{7}{*}{\textbf{Cora}}
& \multicolumn{1}{c|}{\textbf{0}} &   81.46±1.04    &   78.88±0.67   &  77.64±2.48     &   \textbf{83.26±0.79}  & 81.35±1.15      & 81.01±0.98 & \underline{82.23±0.93} & 81.80±1.27&82.14±0.84&82.02±1.00\\  
& \multicolumn{1}{c|}{\textbf{1}} &       75.51±1.31  &74.27±0.93    &     78.09±1.61 &  77.53±1.59   &    78.88±1.57         &  78.31±1.50 & 80.12±1.33&79.37±1.17&\underline{80.47±0.98}&\textbf{81.34±1.43} \\
& \multicolumn{1}{c|}{\textbf{2}}     &     71.01±1.57
    &70.11±1.25 &  76.97±1.15  &  74.16±1.00 & 75.84±1.35   &  76.52±0.98 & 78.78±0.90 &  78.12±1.09&\underline{79.44±1.15}&\textbf{81.46±1.03} \\ 
    % \hline
    % \midrule
% \multirow{3}{*}{\textbf{LightGCN-S}}
& \multicolumn{1}{c|}{\textbf{3}}    &   67.87±1.76  &   68.09±1.03  &  71.46±2.02    &   67.53±0.79   &      69.21±1.35  &  73.60±0.95  & 77.02±1.06&76.83±1.35&\underline{77.61±1.04}&\textbf{80.78±0.93} \\  
& \multicolumn{1}{c|}{\textbf{4}} &       63.93±1.98  &64.94±1.73    &  70.45±3.30 &  65.06±2.38 &   65.96±0.88    & 72.69±0.93 & 75.31±1.27&75.03±0.97&\underline{75.93±1.17}&\textbf{79.77±1.01} \\
                                 & \multicolumn{1}{c|}{\textbf{5}}     &    57.64±1.88
    &63.15±1.73    &   64.04±3.41  & 57.19±2.63   &   61.91±1.46    & 68.65±1.35& 72.17±1.58&72.66±1.18&\underline{74.49±0.97}&\textbf{78.65±1.74} \\ 
    \cmidrule{2-12}
    & \multicolumn{1}{c|}{\textbf{Mean}}     &    69.72
    &69.91    &   73.11  & 70.79   &  72.19    & 75.13& 77.61& 77.30 & \underline{78.35} &\textbf{80.67} \\
    % \hline
    \midrule
\multirow{7}{*}{\textbf{Citeseer}}
& \multicolumn{1}{c|}{\textbf{0}} &   83.12±1.12    &   82.71±0.83   &  78.02±1.35    &  83.23±0.96  &     83.54±0.78     & \underline{85.00±1.91}& \textbf{85.35±2.20} &84.56±1.23&84.74±0.99&83.03±1.05\\
& \multicolumn{1}{c|}{\textbf{1}} &    76.25±1.38  &77.50±2.60   &  78.12±0.81 & 81.25±1.80    &  79.69±1.82        &  \textbf{84.47±1.73}&\underline{82.66±1.64} & 82.38±1.07 &82.51±1.38 &82.50±1.21 \\
& \multicolumn{1}{c|}{\textbf{2}}     &     60.10±4.52
    &65.62±2.13    &     69.90±2.35  &  69.68±5.86  &  74.17±0.78      &  70.83±4.42 &81.15±2.51&80.84±1.55&\underline{81.33±1.01}&\textbf{81.25±1.47} \\ 
    % \hline
    % \midrule
% \multirow{3}{*}{\textbf{LightGCN-S}}
& \multicolumn{1}{c|}{\textbf{3}}    &    54.48±1.92   & 64.79±3.01   &  70.00±1.21    &  58.64±2.47  &   73.65±1.68 &  64.69±3.42&\underline{80.59±1.28} & 77.33±1.34 & 80.36±1.23 &\textbf{81.87±1.06} \\  
& \multicolumn{1}{c|}{\textbf{4}} &       48.44±2.43  &58.12±1.67   &     62.08±3.13 &   53.23±1.35 &   71.77±2.16    & 61.04±2.21 &79.28±1.80 & 76.14±1.28 & \underline{79.67±1.17} &\textbf{82.60±1.61}\\
                                 & \multicolumn{1}{c|}{\textbf{5}}     &    43.33±2.95
    &52.50±1.99   &   57.40±4.26   &  47.50±4.57  &  63.44±3.00  &  56.04±4.57 
  & 78.19±1.74 &74.87±0.88&\underline{78.51±1.39}&\textbf{81.46±1.02} \\  
    \cmidrule{2-12}
    & \multicolumn{1}{c|}{\textbf{Mean}}     &    60.95
    &66.87    &   69.25  & 65.59   &  74.27     &70.35 & \underline{81.20} & 79.35 &81.19&\textbf{82.12} \\
            \midrule

%                 \multirow{7}{*}{\textbf{Chameleon}}
% & \multicolumn{1}{c|}{\textbf{0}} &   \textbf{59.20±0.42}    &   45.61±1.96   &  50.57±1.05   &   58.64±1.01    &    54.34±0.58      &   58.53±1.11&      \underline{59.00±0.74}    &  51.11±1.68 &  58.01±0.86\\  
% & \multicolumn{1}{c|}{\textbf{1}} &       54.58±0.70  &  43.74±1.08    &     49.84±0.85 &  53.93±0.82   &   48.77±1.02   &  56.10±0.48 &    \underline{56.56±0.60}        & 48.93±0.21 &  \textbf{58.45±0.90} \\
% & \multicolumn{1}{c|}{\textbf{2}}     &    50.62±0.87
%     &41.44±1.20   &   49.35±0.82  &  48.67±1.41  &   42.30±1.20   &      53.15±1.06& \underline{54.50±0.87}    &  47.53±2.17  &\textbf{56.60±0.69} \\ 
%     % \hline
%     % \midrule
% % \multirow{3}{*}{\textbf{LightGCN-S}}
% & \multicolumn{1}{c|}{\textbf{3}}    &      47.20±0.67    &  40.07±1.20  &  48.23±0.79    & 43.64±1.53    &   38.75±1.11   & 47.58±1.44 &       \underline{51.60±1.01} & 47.33±0.88  &  \textbf{55.48±0.97} \\  
% & \multicolumn{1}{c|}{\textbf{4}} &       42.84±1.48  & 39.96±1.19    &     47.40±0.78 &    40.23±1.34 &   36.44±1.10      & 44.51±1.25 &  \underline{48.66±1.76 }  & 46.96±1.49 &  \textbf{54.83±0.48} \\
%                                  & \multicolumn{1}{c|}{\textbf{5}}     &    40.63±2.19
%     &37.99±1.27   &   45.74±0.94   &  37.06±1.92  &   35.67±1.41    &  41.45±1.73  &   45.88±2.71    &  \underline{46.52±2.49}& \textbf{54.76±0.81}\\ 
%     \cmidrule{2-11}
%     & \multicolumn{1}{c|}{\textbf{Mean}}     &    49.18
%     &41.47    &   48.52  & 47.03   &    42.71  & 50.22  &   \underline{52.7}    & 48.06   &\textbf{56.34} \\
 \multirow{7}{*}{\textbf{Pubmed}}
& \multicolumn{1}{c|}{\textbf{0}} &   89.12±0.36    &   87.72±0.74   &  84.23±0.89   &   89.37±0.74      &      \underline{89.14±0.67}    &  89.04±0.47 &\textbf{89.19±0.83} &89.07±0.82&OOM&89.13±0.66*\\  
& \multicolumn{1}{c|}{\textbf{1}} &       87.73±0.47  &  86.88±0.51    &    83.25±0.61 &  87.21±0.83  &    87.96±0.53        & 87.13±0.62 &\underline{88.33±1.14}
 &86.69±0.61&OOM&\textbf{88.45±0.78}* \\
& \multicolumn{1}{c|}{\textbf{2}}     &    83.87±0.61
    &86.06±0.69   &   82.18±0.78  &  85.61±0.93   & 87.24±0.77    &  86.48±0.57  &\underline{87.91±1.30}& 85.73±0.92
 &OOM&\textbf{88.01±0.79}* \\ 
    % \hline
    % \midrule
% \multirow{3}{*}{\textbf{LightGCN-S}}
& \multicolumn{1}{c|}{\textbf{3}}    &      80.14±0.58    &  85.03±0.37  &  81.09±0.84    & 84.23±1.14   &       86.39±0.71 & 85.51±0.61  &\underline{87.20±1.49} & 83.12±0.88 &OOM &\textbf{87.68±0.51}* \\  
& \multicolumn{1}{c|}{\textbf{4}} &       75.39±0.75  & 83.94±0.56    &     80.23±0.67 &    80.14±0.83 &  85.64±0.46  & 84.87±0.48   &\underline{86.53±0.91}&81.71±0.87&OOM&\textbf{87.23±0.81}* \\
                                 & \multicolumn{1}{c|}{\textbf{5}}     &    65.81±0.47
    &83.53±0.77   &   77.35±1.02   &  71.86±0.91 &   85.06±0.78    &  84.26±0.59 &\underline{86.11±1.32} &80.48±0.81&OOM&\textbf{86.94±0.67}*\\ 
    \cmidrule{2-12}
    & \multicolumn{1}{c|}{\textbf{Mean}}     &    80.34
    &85.53    &   81.39  & 83.07   &    86.91    & 86.22   &\underline{87.55} &84.47&OOM&\textbf{87.91}* \\
    \bottomrule
\end{tabular}}
% \vspace{-2mm}
\end{table*}

\begin{figure*}[!ht]
%\renewcommand{\subfigcapskip}{-4pt} c
%\renewcommand{\subfigbottomskip}{0pt}
% \vskip -0.2in
\centering
{\subfigure[Cora]
{\includegraphics[width=0.32\linewidth]{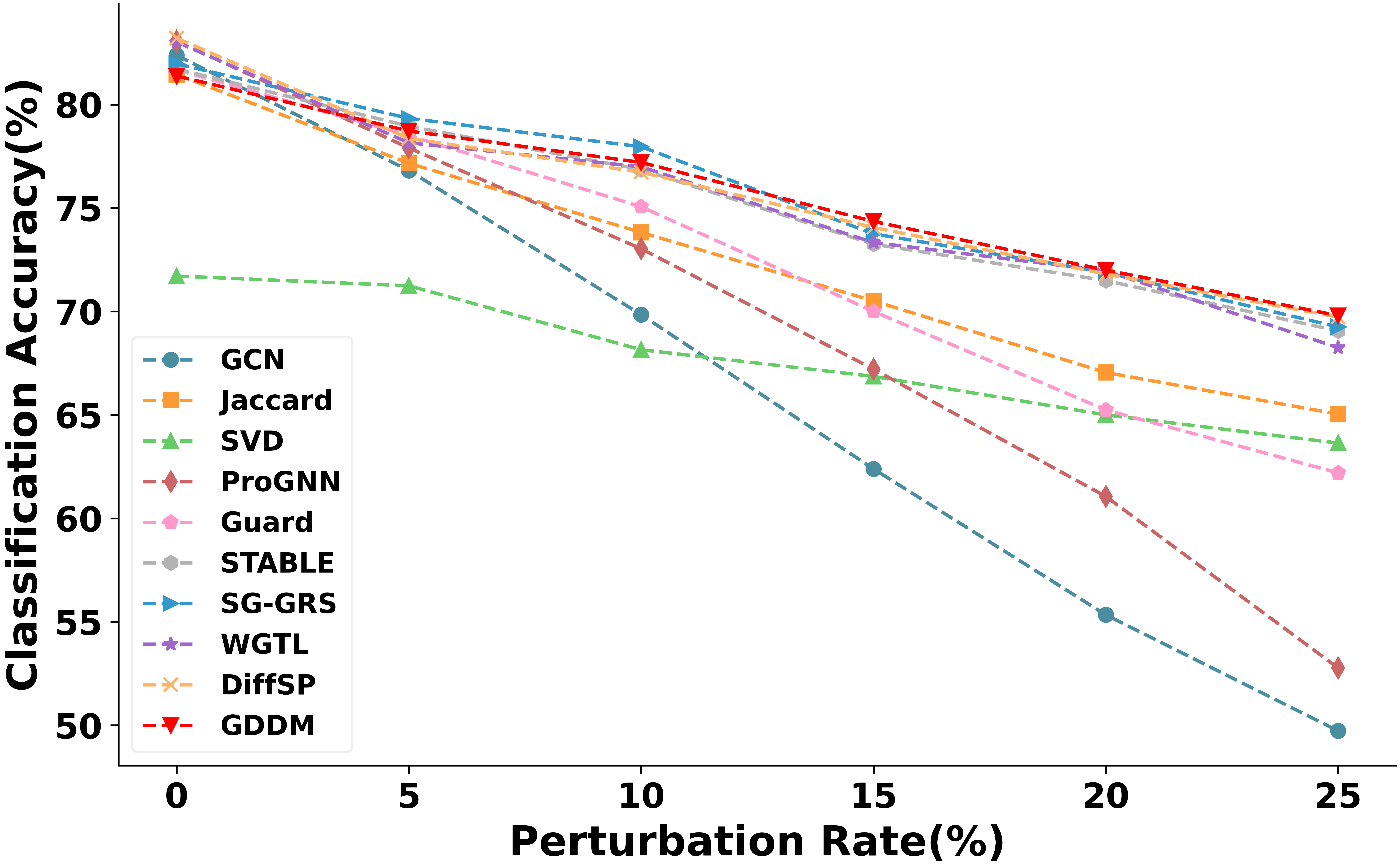}}
{\subfigure[Citeseer]
{\includegraphics[width=0.32\linewidth]{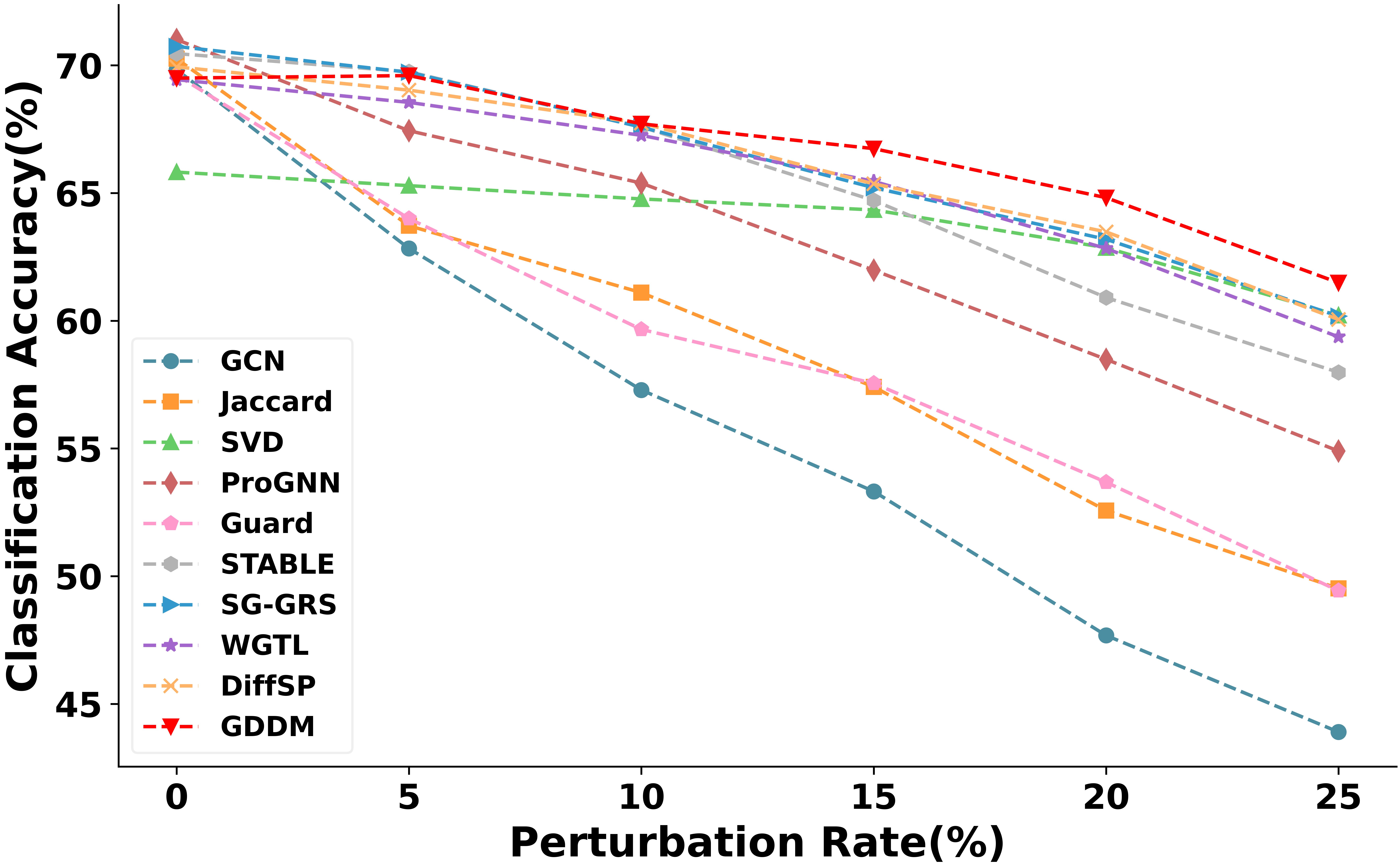}}}
{\subfigure[Pubmed]
{\includegraphics[width=0.32\linewidth]{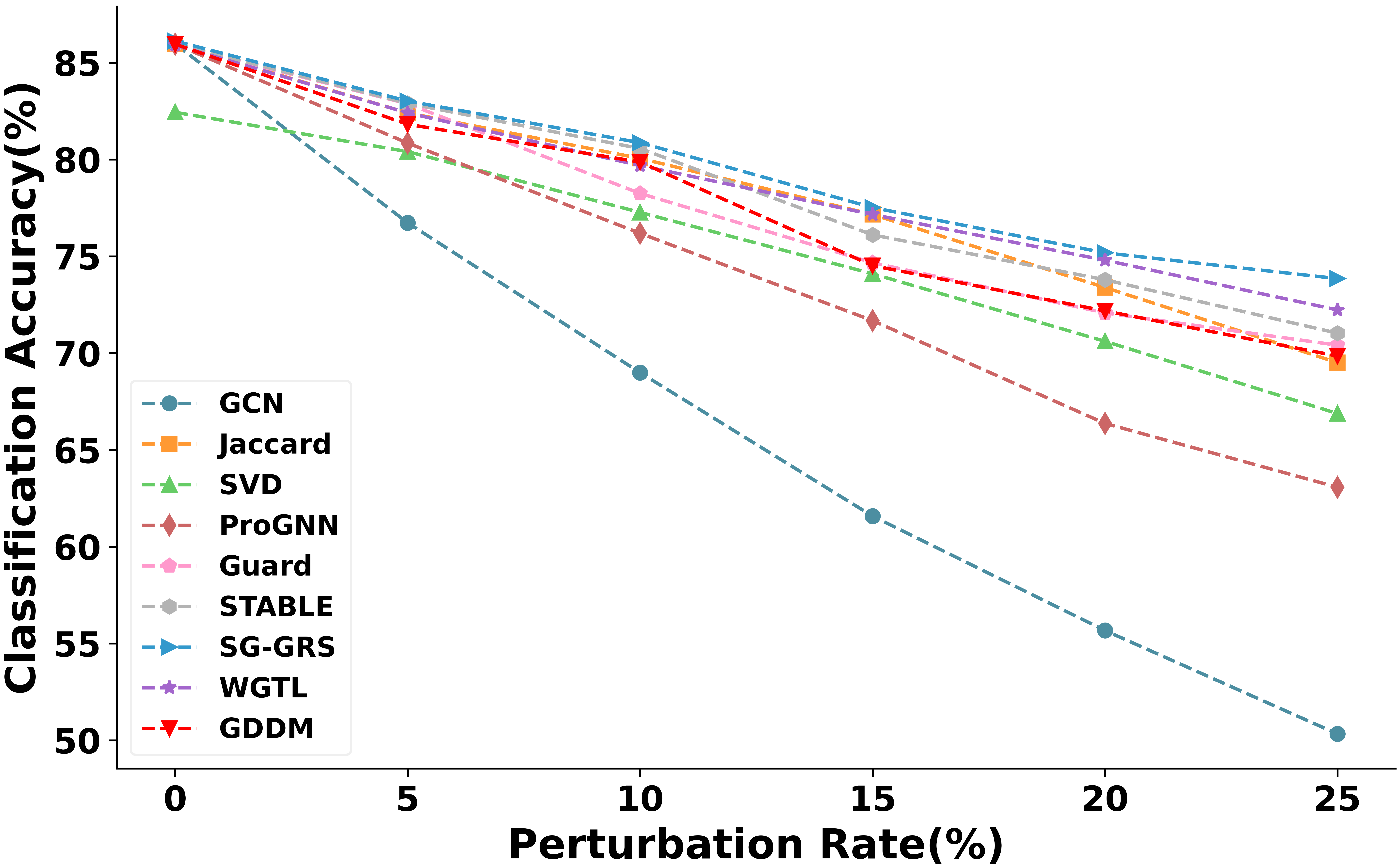}}}}%

% \vskip -0.2in
% \vspace{-4mm}
\Description{Node classification performance}
\caption{Node classification performance (Accuracy±Std) under non-targeted attack (GraD).}\label{fig:node_class_per_under_non_tar_att}
% \vskip -0.2in
% \vspace{-4mm}
\end{figure*}

% \subsubsection{Baselines.}
% We compare GDDM with representative and state-of-the-art defense methods to evaluate its robustness. These baseline models include classical GNNs method \textbf{GCN}~\cite{kipf2016semi}, improved GCNs methods
% \textbf{ProGNN}~\cite{jin2020graph}, 
% \textbf{STABLE}~\cite{li2022reliable}, \textbf{SG-GRS}~\cite{in2024self},
% \textbf{WGT-L}~\cite{arafat2025witnesses}, graph purification methods \textbf{Jaccard}~\cite{wu2019adversarial}, \textbf{SVD}~\cite{entezari2020all}, \textbf{Guard}~\cite{li2023guard}, and graph diffusion-based model \textbf{DiffSP}~\cite{luo2025robust}. For the experiments, we implement a non-targeted structural attack \textbf{GraD}~\cite{liu2023towards}, and a targeted structural attack \textbf{Nettack}~\cite{zugner2018adversarial}. Detailed descriptions them are provided in Appendix~\ref{appendix:baseline}.

\subsubsection{Experimental Details.}
We compare GDDM with representative and state-of-the-art defense methods including classical GNNs method~\cite{kipf2016semi}, improved GCNs methods~\cite{jin2020graph,li2022reliable,in2024self,arafat2025witnesses}, graph purification methods~\cite{wu2019adversarial,entezari2020all,li2023guard} and  graph diffusion-based method~\cite{luo2025robust} to evaluate its robustness.
We mainly consider \textbf{poisoning attacks}(non-targeted structure attack~\cite{liu2023towards} and targeted structure attack~\cite{zugner2018adversarial}) in this work. 
Our experiments follow the \textbf{transductive setting}, where the training and test nodes are available during training—an approach widely adopted in adversarial defense for node classification.
We provide descriptions of the baselines, implementation details in Appendix~\ref{appendix:baseline} and Appendix~\ref{appendix:implementation_details}, respectively.
% We then generate attacks on each graph based on the perturbation number$/$rate.
% % and we ensure that all hyper-parameters in the attack methods align with the original authors’ implementations.
% During the model training phase, we primarily adjust the diffusion steps, with a range of $\{64, 128, 256, 512\}$.
% During the model inference phase, we primarily adjust the expected size of the generated graph, ranging from 70\% to 95\%.
% Performance is reported as the average accuracy with standard deviation, based on 10 runs on clean, perturbed or denoised graphs.
% We closely follow the default setting of the baselines for a fair comparison.
% Our experiments follow the \textbf{transductive setting}, where the training and test nodes are available during training—an approach widely adopted in adversarial defense for node classification.
% And we provide a guide for fast hyper-parameter tuning in Appendix~\ref{appendix:guide_for_Hpt}.

% \vspace{-2mm}
\subsection{Robustness Against Poisoning Attacks}\label{sec:rob_eva}
\subsubsection{Performance Against Targeted Attack.} \label{sec:per_against_tar}
\textbf{Q: Does GDDM outperform state-of-the-art graph defense models across targeted attacks while maintaining strong scalability?} The answer is \textbf{Yes}, since GDDM outperforms other methods on the majority of datasets and can successfully transfer from small to large datasets for effective defense.
Table~\ref{tab:performance_com_under_tar_att} presents the performance on three datasets against targeted attack (Nettack~\cite{zugner2018adversarial}). 
We set the perturbation number for the targeted node between 1 and 5 with the step of 1.
Following prior work~\cite{jin2020graph}, we selected nodes with degrees larger than 10 in the test set as target nodes for the targeted attacks.
% We present a comparison in the Appendix~\ref{appendix:efficiency_comp} between GDDM and other baseline methods on a larger dataset to validate its scalability further.
The top two results are highlighted in bold and underlined. 
% We set the perturbation rate between 0\% and 25\% with the step of 5\% under non-targeted attack. 
From this main experiment, we make the following observations:

\begin{itemize}
    \item The performance of all the methods declines as the perturbation number increases, demonstrating the effectiveness of the attack method. 
    In most cases, GCN shows the largest performance drop, proving that its ability to defend against adversarial attacks is particularly weak.
    \item Jaccard and SVD are feature-based edge pruning methods that show lower sensitivity to perturbations compared to other baselines. However, they often fail to fully remove adversarial edges and may mistakenly delete clean edges, resulting in performance degradation on clean graphs. Guard uses a scoring function based on gradients and node degrees, therefore outperforming them in most cases.
    \item The improved GCN methods ProGNN, STABLE, SG-GRS and WGTL perform well with low perturbations but ProGNN and STABLE suffer a sharp decline as perturbations increase, likely due to unreliable structure learning in heavily contaminated graphs. STABLE outperforms ProGNN by adding a greater number of reliable edges.
    SG-GRS and WGTL enhance their robustness under strong adversarial attacks by incorporating additional constraints.
    % GDDM outperforms all by leveraging the denoising capability of the graph diffusion model and a tailored strategy for targeted attacks.
    \item Graph diffusion-based model DiffSP models both the attack and defense processes using a diffusion-based framework, which contributes to its strong robustness on most datasets. However, despite its effectiveness, DiffSP suffers from poor scalability and fails to run on larger datasets like Pubmed due to high training cost.
    \item Experimental results on the Pubmed dataset show that GDDM effectively defends against adversarial attacks when transferred from a small to a large dataset, highlighting its strong scalability and providing a viable approach for deploying graph diffusion models on large-scale datasets.
    % The results of GDDM on the Pubmed dataset demonstrate that directly transferring a model trained on a smaller dataset to a larger one still effectively defends against adversarial attacks. This highlights the strong scalability of GDDM and provides a promising approach for deploying graph diffusion models on large-scale datasets.
    \item GDDM surpasses baseline methods across most perturbation levels, benefiting from the diffusion model's strong recovery capability in highly disruptive scenarios,  along with our tailored denoising strategies and components that leverage structural and feature information for guidance.
\end{itemize}
% \vspace{-1mm}

\subsubsection{Performance Against Non-targeted Attack.} \textbf{Q: Does GDDM outperform state-of-the-art graph defense models across non-targeted attacks?} The answer is \textbf{Yes}, since GDDM performs well on most datasets in non-targeted attack scenarios.
The performance under non-targeted attack is shown in Fig.~\ref{fig:node_class_per_under_non_tar_att}. We chose the powerful attack method GraD which is an improved version of Metattack~\cite{DBLP:conf/iclr/ZugnerG19} as an attacker to perturb graph structure. As we can see, GDDM is competitive compared to Sota methods, and in most cases, its performance outperforms other baseline methods.
Together with the observations from Section~\ref{sec:per_against_tar}, we can conclude that GDDM is capable of effectively defending against both targeted and non-targeted attacks. 

\begin{table}[!ht]
\centering
% \vspace{-3mm}
\caption{Results of denoising strategies for a targeted attack.}
% \vspace{-3mm}
\label{tab:com_nt_and_t}
\scalebox{0.9}{
\begin{tabular}{c|cc|c}
% \specialrule{0.08em}{0pt}{0pt}
\toprule
\multicolumn{1}{c}{\textbf{Datasets}}              & 
\multicolumn{1}{|c|}{\textbf{Ptb Num}} & \multicolumn{1}{c|}{\textbf{GDDM w/o TADS}}& \multicolumn{1}{c}{\textbf{GDDM}}\\
% \hline
\midrule
\multirow{3}{*}{\textbf{Cora}}
& \multicolumn{1}{c|}{\textbf{0}} &   80.80±1.57    &   \textbf{82.02±1.00}\\  
& \multicolumn{1}{c|}{\textbf{1}} &       77.41±0.85  &\textbf{81.34±1.43} \\
& \multicolumn{1}{c|}{\textbf{2}}     &     72.58±1.35
    &\textbf{81.46±1.03}\\ 
& \multicolumn{1}{c|}{\textbf{3}}    &   70.79±1.23  &   \textbf{80.78±0.93 }\\  
& \multicolumn{1}{c|}{\textbf{4}} &       66.06±1.87  &\textbf{79.77±1.01}  \\
                                 & \multicolumn{1}{c|}{\textbf{5}}     &    62.13±2.71
    &\textbf{78.65±1.74}\\ 
    % \cmidrule{2-4}
    % & \multicolumn{1}{c|}{\textbf{Mean}}     &    71.63
    % &\textbf{80.67} \\
    % \hline
    \midrule
    
\multirow{3}{*}{\textbf{Citeseer}}
& \multicolumn{1}{c|}{\textbf{0}} &   82.02±1.48    &   \textbf{83.03±1.05} \\  
& \multicolumn{1}{c|}{\textbf{1}} &    81.25±1.55  &\textbf{82.50±1.21} \\
& \multicolumn{1}{c|}{\textbf{2}}     &     78.75±1.87
    &\textbf{81.25±1.47} \\ 
& \multicolumn{1}{c|}{\textbf{3}}    &    78.02±1.51   & \textbf{81.87±1.06} \\  
& \multicolumn{1}{c|}{\textbf{4}} &       76.88±2.02  & \textbf{81.60±1.61} \\
                                 & \multicolumn{1}{c|}{\textbf{5}}     &    79.06±1.58
    &\textbf{81.46±1.02}  \\ 
    % \cmidrule{2-4}
    % & \multicolumn{1}{c|}{\textbf{Mean}}     &    79.33
    % &\textbf{82.12}  \\
            \midrule
%     \multirow{7}{*}{\textbf{Chameleon}}
% & \multicolumn{1}{c|}{\textbf{0}} &   57.88±0.52    &   \textbf{58.01±0.86} \\  
% & \multicolumn{1}{c|}{\textbf{1}} &       57.51±0.93  &  \textbf{58.45±0.90} \\
% & \multicolumn{1}{c|}{\textbf{2}}     &    55.69±0.98
%     &\textbf{56.60±0.69} \\ 
% & \multicolumn{1}{c|}{\textbf{3}}    &      55.31±1.16    &  \textbf{55.48±0.97} \\  
% & \multicolumn{1}{c|}{\textbf{4}} &       52.26±0.80  & \textbf{54.83±0.48} \\
%                                  & \multicolumn{1}{c|}{\textbf{5}}     &    53.93±1.25
%     &\textbf{54.76±0.81}  \\ 
%     \cmidrule{2-4}
%     & \multicolumn{1}{c|}{\textbf{Mean}}     &    55.43
%     &\textbf{56.36} \\
    \multirow{3}{*}{\textbf{Pubmed}}
& \multicolumn{1}{c|}{\textbf{0}} &   86.78±0.62    &   \textbf{89.13±0.66} \\  
& \multicolumn{1}{c|}{\textbf{1}} &       85.63±0.81  &  \textbf{88.45±0.78} \\
& \multicolumn{1}{c|}{\textbf{2}}     &    84.43±0.72
    &\textbf{88.01±0.79} \\ 
& \multicolumn{1}{c|}{\textbf{3}}    &      82.13±0.65    &  \textbf{87.68±0.51} \\  
& \multicolumn{1}{c|}{\textbf{4}} &       80.07±0.74  & \textbf{87.23±0.81} \\
                                 & \multicolumn{1}{c|}{\textbf{5}}     &    77.42±0.89
    &\textbf{86.94±0.67}  \\ 
    % \cmidrule{2-4}
    % & \multicolumn{1}{c|}{\textbf{Mean}}     &    82.74
    % &\textbf{87.91} \\
    % \hline\specialrule{0.08em}{0pt}{0pt}
    \bottomrule
\end{tabular}}
% \vspace{-4mm}
% \vskip -0.1in
\end{table}
% We provide an evaluation of GDDM under other attack scenarios, with results presented in the Appendix~\ref{appendix:node_inject_attack}.

\begin{table*}[!ht]
\centering
% \vskip -0.1in
\caption{Ablation experiments under non-targeted attack (GraD) on Cora and Citeseer.}
% \vspace{-3mm}
% \vskip -0.1in
\label{tab:ablation_exp}
\scalebox{0.8}{
\begin{tabular}{c|ccccc|ccccc}
% \specialrule{0.08em}{0pt}{0pt}
\toprule
\multicolumn{1}{c|}{\textbf{Dataset}}              & 
\multicolumn{5}{c|}{\textbf{Cora}} & \multicolumn{5}{c}{\textbf{Citeseer}}\\
\midrule
\multicolumn{1}{c|}{\textbf{Ptb Rate}} &
\multicolumn{1}{c}{\textbf{5\%}}&
\multicolumn{1}{c}{\textbf{10\%}}&
\multicolumn{1}{c}{\textbf{15\%}}&
\multicolumn{1}{c}{\textbf{20\%}}&
\multicolumn{1}{c|}{\textbf{25\%}}&
\multicolumn{1}{c}{\textbf{5\%}}&
\multicolumn{1}{c}{\textbf{10\%}}&
\multicolumn{1}{c}{\textbf{15\%}}&
\multicolumn{1}{c}{\textbf{20\%}}&
\multicolumn{1}{c}{\textbf{25\%}}\\
% \hline
\midrule
\multicolumn{1}{c|}{\textbf{GDDM w/o GSDR}}
& 32.49±0.47 &   32.49±0.47    &   32.49±0.47   &  32.49±0.47     &   32.49±0.47     &    33.19±0.62       & 33.19±0.62 & 33.19±0.62      & 33.19±0.62 & 33.19±0.62\\  
\multicolumn{1}{c|}{\textbf{GDDM w/o NFCR}}
& 77.76±0.45 &   75.65±0.47    &   72.51±0.32   &  69.77±0.50     &   66.85±0.28     &    68.75±0.29       &66.76±0.52& 65.67±0.47      & 64.01±0.72 & 60.68±0.38\\  
% \multicolumn{1}{c|}{\textbf{GDDM w/o ND}}&        78.31±0.57  & 76.49±0.47    &     71.86±0.59 &  70.73±0.47    &   68.19±0.28       &  69.12±0.26 &    67.49±0.34         &  66.24±0.52 & 64.35±0.62 & 61.00±0.57 \\
\multicolumn{1}{c|}{\textbf{GDDM}}& \textbf{78.72±0.54}     &     \textbf{77.20±0.34}
    &\textbf{73.35±0.40} &  \textbf{72.00±0.50}  &  \textbf{69.80±0.32}  &     \textbf{69.60±0.35}    &   \textbf{67.71±0.43}& \textbf{66.74±0.40}   &  \textbf{64.82±0.59} & \textbf{61.49±0.47} \\ 
    \midrule
    \multicolumn{1}{c|}{\textbf{Ptb Num}} &
\multicolumn{1}{c}{\textbf{1}}&
\multicolumn{1}{c}{\textbf{2}}&
\multicolumn{1}{c}{\textbf{3}}&
\multicolumn{1}{c}{\textbf{4}}&
\multicolumn{1}{c|}{\textbf{5}}&
\multicolumn{1}{c}{\textbf{1}}&
\multicolumn{1}{c}{\textbf{2}}&
\multicolumn{1}{c}{\textbf{3}}&
\multicolumn{1}{c}{\textbf{4}}&
\multicolumn{1}{c}{\textbf{5}}\\
    \midrule
    \multicolumn{1}{c|}{\textbf{GDDM w/o GSDR}}
& 42.41±3.01 &  42.41±3.01    &   42.41±3.01   &  42.41±3.01     &   42.41±3.01     &    56.90±4.37      & 56.90±4.37& 56.90±4.37      &56.90±4.37 & 56.90±4.37\\ 
    \multicolumn{1}{c|}{\textbf{GDDM w/o NFCR}}
& 81.12±1.21 &   81.01±1.06    &   80.56±1.01   &  79.55±1.40     &   78.09±1.44     &    81.45±1.30      & 81.15±1.43& 81.25±1.04      & 82.08±1.38 & 81.04±1.12\\  
% \multicolumn{1}{c|}{\textbf{GDDM w/o ND}}&        81.24±1.40  & 81.23±1.13    &     80.44±0.90 &  79.33±1.35    &   78.42±0.84       &   81.67±0.95 &    80.83±1.88         &  80.52±1.24 & 81.88±1.16 & 80.70±1.63 \\
\multicolumn{1}{c|}{\textbf{GDDM}}& \textbf{81.34±1.43}     &     \textbf{81.46±1.03}
    &\textbf{80.78±0.93} &  \textbf{79.77±1.01}  &  \textbf{78.65±1.74}  &     \textbf{82.50±1.21}    &   \textbf{81.25±1.47}& \textbf{81.87±1.06}   &  \textbf{82.60±1.61} & \textbf{81.46±1.02} \\ 
    \bottomrule
\end{tabular}}
% \vspace{-4mm}
% \vskip -0.1in
\end{table*}

\begin{table*}[!ht]
% \label{table3*}
\centering
% \vskip -0.1in
\renewcommand\arraystretch{1}
\caption{Average training time per epoch (sec/epoch) on the Cora dataset.}
\label{tab:efficiency_comparison}
\scalebox{1}{
\begin{tabular}{c|ccccccccc}
    \toprule
    Model & SVD & Jaccard&Guard&ProGNN&STABLE&WGTL&SG-GSR&DiffSP&  GDDM\\
    \midrule
    sec/epoch& 0.013 & 0.009&0.005&3.57&0.078&3.12&0.057&23.64&2.45\\
    \bottomrule
\end{tabular}}
\end{table*}

\subsection{Impact of Attack-specific Denoising}
\textbf{Q: Do our designed denoising strategies enhance GDDM's ability to protect specific nodes under targeted attack?} The answer is \textbf{Yes}, since the Tailored Attack-specific Denoising Strategies (TADS) enhances GDDM's defense effectiveness in targeted attack scenarios.
To validate the effectiveness of our proposed TADS for targeted attacks, we conducted separate experiments in a targeted attack scenario using both GDDM's denoising strategies designed for targeted attacks and non-targeted attacks.
The experimental results are shown in Table~\ref{tab:com_nt_and_t}, where GDDM represents the performance of GDDM using the denoising strategy designed for targeted attacks in a targeted attack scenario, and GDDM w/o TADS represents the performance of GDDM using the denoising strategy for non-targeted attacks in the same targeted attack scenario. 
The results indicate that the performance of GDDM in the targeted attack scenario significantly improves when using the TADS, demonstrating that GDDM uses TADS to achieve targeted protection by focusing on the attacked nodes.

% \vspace{-2mm}
\subsection{Ablation Study}
\textbf{Q: Do the additional components improve the model's defense effectiveness against various attack methods?} The answer is \textbf{Yes}, the removal of different modules negatively impacts GDDM to varying degrees.
To gain a deeper understanding of how different components impact the model's performance,
we conduct an ablation study to evaluate the contributions of the various components in GDDM:
\begin{itemize}
    \item \textbf{GDDM w/o NFCR}: the proposed GDDM without the Node Feature-Constrained Regularizer.
    \item \textbf{GDDM w/o GSDR}: the proposed GDDM without the Graph Structure-Driven Refiner.
    % \item \textbf{GDDM w/o FD\&ND\&SG}: GDDM without feature smoothness, node degree guiding and Graph Structure Guiding Module.
\end{itemize}

We use the hyper-parameters that achieve the best performance in our main experiment and report the results of 10 runs in Table~\ref{tab:ablation_exp}.
% The complete results are provided in Appendix~\ref{appendix:chameleon_ablation}.
By comparing the full GDDM model with its ablation versions \textbf{GDDM w/o GSDR}, \textbf{GDDM w/o NFCR}, we observe that the Graph Structure-Driven Refiner plays a crucial role in GDDM. This component guides the generation process using the structure of the attacked graph, ensuring the basic graph fidelity of the denoised graph.
Furthermore, we also observe that both feature smoothness guiding and node degree guiding consist in the Node Feature-Constrained Regularizer future enhance the defense performance. 
The above two experimental phenomena highlight the importance of graph fidelity in defending against adversarial attacks in graph diffusion models.
% Feature smoothness improves the model's performance, while incorporating node degree further boosts it.
% \vspace{-1.5mm}
% \vspace{-2mm}
\subsection{Model Analyses}
% \subsubsection{Flexibility Analysis}
% \textbf{Q: How does the flexibility of GDDM with downstream GNNs, attack scenarios, and larger datasets?} The answer is \textbf{Yes}. 
% First, we validate that GDDM can be integrated with different GNNs to mitigate graph adversarial attacks. 
% Second, beyond structure attacks, we conduct preliminary evaluations under injection attacks~\cite{zou2021tdgia}, showing that GDDM maintains competitive performance even when facing attacks beyond the design assumption.
% Third, we test the scalability of GDDM by transferring it to a larger dataset (Arxiv) and comparing its performance with other baselines. The results show that GDDM can be successfully deployed on larger graphs, highlighting its superior scalability.
% The experimental results and analyses are provided in Appendix~\ref{appendix:flexibility}, Appendix~\ref{appendix:node_inject_attack}, and Appendix~\ref{appendix:large_scale_dataset}, respectively.

% From the experimental results, we observe that GDDM demonstrates a highly significant denoising capability on the attacked graphs across three different GNN classifiers. This indicates that GDDM can be flexibly integrated between upstream attack methods and downstream GNN classifiers, significantly enhancing the robustness of the GNN classifiers.
% The experimental results are shown in Appendix~\ref{appendix:flexibility}.

% \vspace{-1.5mm}
\subsubsection{Parameter Analysis}
\begin{figure}[!t]
% \vspace{-2mm}
\centering
{\subfigure[Cora]
{\includegraphics[width=0.49\linewidth]{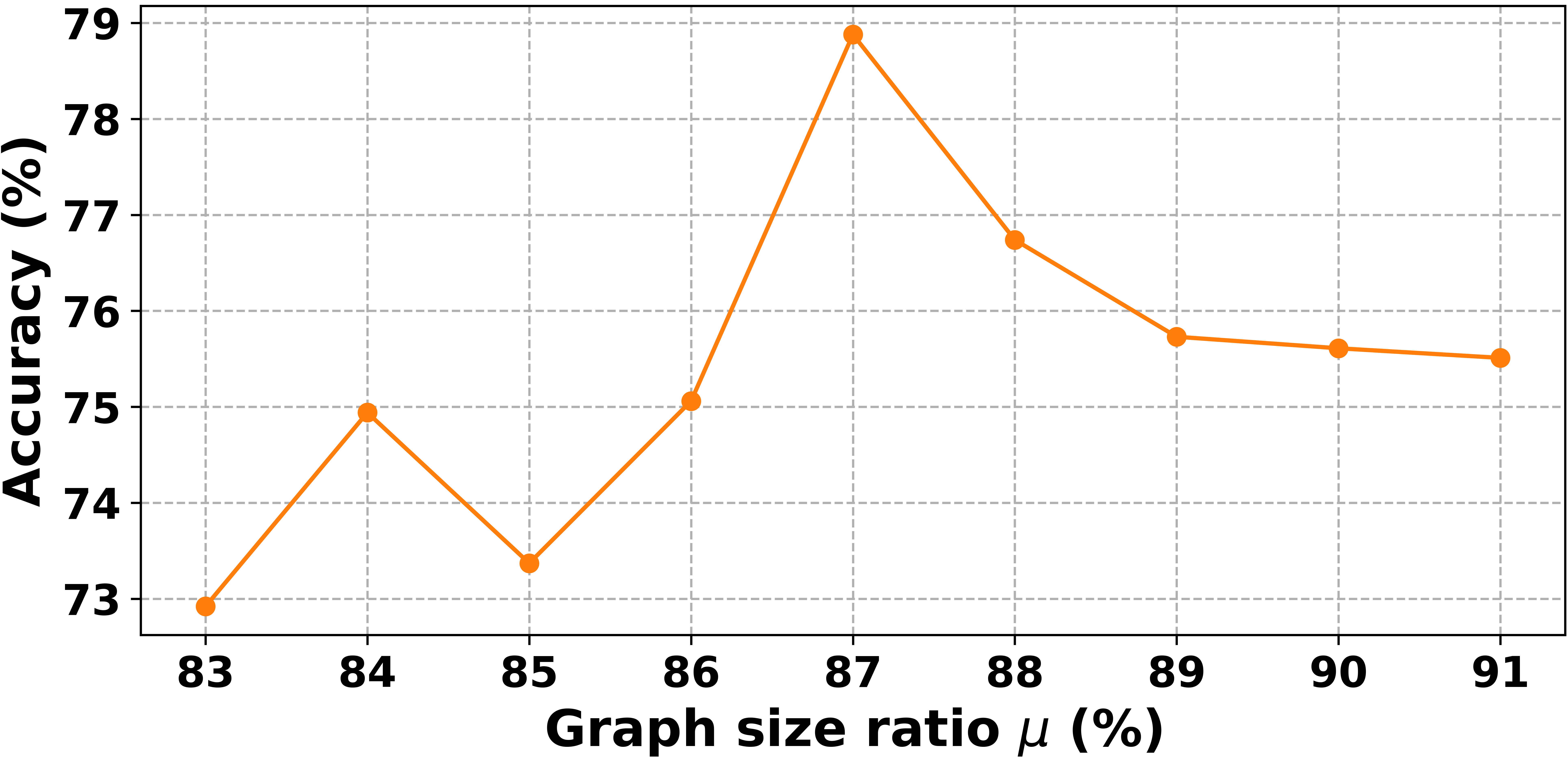}}
{\subfigure[Citeseer]
{\includegraphics[width=0.49\linewidth]{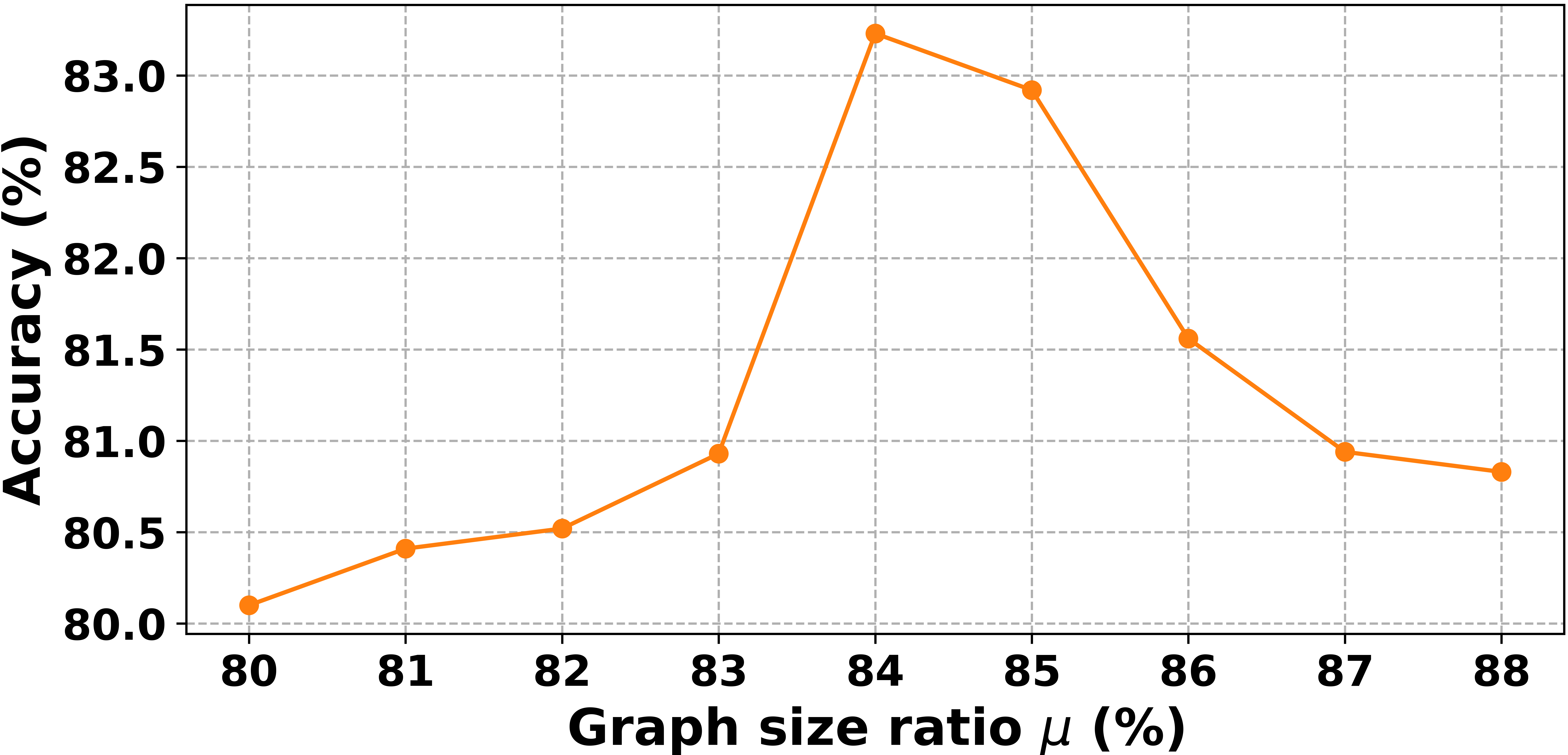}}}}%
% \vspace{-4mm}
\Description{Graph size ratio parameter analysis}
\caption{Graph size ratio parameter analysis.}\label{fig:para_an_mu}
\end{figure}

\textbf{Q: How does the performance of GDDM vary with different hyper-parameters?} Drawing from our experimental observations, the graph size ratio $\mu$ plays a critical role in optimizing performance. 
Accordingly, we systematically vary its value to evaluate its influence on the performance of GDDM.
% From the experimental results, we can observe that as the graph size ratio $\mu$ increases, the model's performance first improves and then declines. 
% This may be because a graph size ratio that is too small can cause the model to remove clean edges from the attacked graph, while a ratio that is too large may lead to insufficient removal of noisy edges.
% Therefore, we select an optimal graph size ratio to achieve the best model performance.
The results are shown in Fig.~\ref{fig:para_an_mu}.
Drawing from our experimental observations, the graph size ratio $\mu$ plays a critical role in optimizing performance. 
Accordingly, we systematically vary its value to evaluate its influence on the performance of GDDM.
% As shown in Fig.~\ref{fig:para_an_mu}, we conduct experiments using the targeted attack method Nettack on Cora and Citeseer datasets with a larger perturbation number (num=5).
It is worth noting that as the graph size ratio $\mu$ increases, the model's performance first improves and then declines. 
This may be because a graph size ratio that is too small can cause the model to remove clean edges from the attacked graph, while a ratio that is too large may lead to insufficient removal of noisy edges.
Therefore, we select an optimal graph size ratio to achieve the best model performance.

\subsubsection{Node Degree Distribution Similarity Analysis}
\textbf{Q: Does node degree distribution similarity significantly affect the transfer performance of GDDM?} The answer is \textbf{Yes}. 
To investigate the impact of node degree distribution similarity on the transfer performance of GDDM, we trained the model on the QM9 dataset, which has low degree distribution similarity with Pubmed, and then transferred it to the Pubmed dataset. The results are shown in Table~\ref{tab:impact_of_node_degree}. We observed that the model trained on QM9 performed significantly worse on Pubmed compared to the model transferred from Cora. This result indicates that the similarity in node degree distributions plays an important role in the transferability of GDDM.

\subsubsection{Complexity Analysis}
\textbf{Q: Does GDDM have lower time complexity than existing graph diffusion models?} The answer is \textbf{Yes}. 
Let $M$ denote the number of edges and $K$ the number of nodes with changing degrees during the reverse process. At each step $t$, the MPNN requires $O(M)$ for node representations, $O(N)$ for predicting $\mathbf{s}^t$, and $O(K^2)$ for link predictions among $K$ nodes. With $K \ll N$ in most cases, and $T$ diffusion steps, the total complexity is $O(T \cdot \text{max}(K^2, M))$. In contrast, prior models require $O(T \cdot N^2)$ for $O(N^2)$ link predictions per step.
Additionally, since our model does not require retraining on large datasets, its time overhead in real-world scenarios is lower than other graph diffusion models.

To further validate the complexity advantage of GDDM over existing graph diffusion models, we compare the per-epoch training time of all baseline methods on the Cora dataset. The results are shown in Table~\ref{tab:efficiency_comparison}. All methods were trained on the same data splits to ensure comparable training sizes across models. As shown in Table~\ref{tab:efficiency_comparison}, GDDM incurs a higher per-epoch training cost (2.45 sec/epoch). However, its ability to transfer across datasets—such as from Cora to Pubmed—without retraining helps mitigate this overhead, making it a practically scalable solution.

\begin{table}[!ht]
% \label{table3*}
\centering
% \vskip -0.1in
\renewcommand\arraystretch{1}
\caption{Impact of node degree similarity on GDDM.}
\label{tab:impact_of_node_degree}
\scalebox{1}{
\begin{tabular}{c|cc}
    \toprule
    Model & Source→Target & Accuracy (\%)\\
    \midrule
    GDDM & Cora→Pubmed & \textbf{86.94±0.67}\\ 
    GDDM & QM9→Pubmed & 69.73±1.51\\ 
    % ProGNN& - & 71.86±0.91\\
    % Jaccard+GCN& - & 83.53±0.77\\
    % SVD+GCN& - & 77.35±1.02\\
    % Guard+GCN& - & 85.06±0.78\\
    % STABLE& - & 84.26±0.59\\
    % SG-GRS& - & 86.11±1.32\\
    % GCN& - & 65.81±0.47\\
    \bottomrule
\end{tabular}}
\end{table}

\section{Conclusion}
Graph Neural Networks (GNNs) are highly sensitive to adversarial attacks on graph structures. 
To effectively defend against different types of adversarial attacks that can poison GNNs, we propose a novel graph purification method GDDM in this work. 
% This method leverages the powerful denoising capabilities of diffusion models, supplemented by multiple key components and denoising strategies to purify the attacked graphs, thereby effectively reducing the impact of different types of adversarial attacks on GNNs.
This method leverages the powerful denoising capabilities of diffusion models, supplemented by multiple key components and tailored denoising strategies to purify attacked graphs, effectively reducing the impact of various adversarial attacks on GNNs.
Additionally, this method can leverage the similarity in node degree distributions across graph datasets to enable seamless transfer between similar datasets without retraining, achieving strong scalability.
Our experiments demonstrate that GDDM is highly competitive in different types of attack scenarios.
% outperforming SOTA baseline methods in most cases while maintaining strong scalability. 
% It significantly enhances the performance of Graph Neural Networks under adversarial attack scenario.
\section{Acknowledgments}
This work was supported by a grant from the National Natural Science Foundation of China under grants (No.62372211), and the Science and Technology Development Program of Jilin Province (No.20250102216JC).

\clearpage
\bibliographystyle{ACM-Reference-Format}
\bibliography{sample-base}

\clearpage
\appendix

\appendix

\section{Detailed Implementations}
\subsection{Derivation Process of the Posterior}\label{appendix:derivation_process_of_posterior}
The further derivation of Eq.~\ref{eq:given_a0_at_st_to_at-1} are shown as follows: 
\begin{align}
    q(\mathbf{A}^{t-1}|\mathbf{A}^{t},\mathbf{s}^{t},\mathbf{A}^{0})&=\frac{q(\mathbf{A}^{t-1},\mathbf{A}^{t},\mathbf{s}^{t}|\mathbf{A}^{0})}{q(\mathbf{A}^{t},\mathbf{s}^{t}|\mathbf{A}^{0})} \nonumber \\
    &=\frac{q(\mathbf{A}^{t}|\mathbf{A}^{t-1},\mathbf{A}^{0})q(\mathbf{s}^{t}|\mathbf{A}^{0},\mathbf{A}^{t},\mathbf{A}^{t-1})q(\mathbf{A}^{t-1}|\mathbf{A}^{0})}{q(\mathbf{A}^{t},\mathbf{s}^{t}|\mathbf{A}^{0})} \nonumber\\
    &=\frac{q(\mathbf{A}^{t}|\mathbf{A}^{t-1})q(\mathbf{s}^{t}|\mathbf{A}^{t},\mathbf{A}^{t-1})q(\mathbf{A}^{t-1}|\mathbf{A}^{0})}{q(\mathbf{A}^{t}|\mathbf{A}^{0})q(\mathbf{s}^{t}|\mathbf{A}^{t},\mathbf{A}^{0})}.
\end{align}

\subsection{The Detailed Derivation of Eq.~\ref{eq:pred_st_given_d0_and_dt}} \label{appendix:derivation_of_eq7}
The content of Eq.~\ref{eq:pred_st_given_d0_and_dt} is shown below:
\begin{align}
q(\mathbf{s}^{t}|\mathbf{d}^{t},\mathbf{d}^{0})&=\prod_{i=1}^{N}q(\mathbf{s}^{t}_{i}|\mathbf{d}^{t}_{i},\mathbf{d}^{0}_{i}),\\
q(\mathbf{s}^{t}_{i}|\mathbf{d}^{t}_{i},\mathbf{d}^{0}_{i})&=\mathcal{B}(\mathbf{s}^{t}_{i};1-(1-\frac{\beta_{t}\bar{\alpha}_{t-1}}{1-\bar{\alpha}_{t-1}})^{\mathbf{d}^{0}_{i}-\mathbf{d}^{t}_{i}}),\nonumber
\end{align}

For the second line of Equation 7, we have added the detailed derivation process as follows:

First, the degree distribution in the forward diffusion process of the graph takes the following form:
\begin{align}
q(\mathbf{d}^t|\mathbf{d}^0)&=\prod \limits_{i=1}^N q(\mathbf{d}^t_i|\mathbf{d}^0_i),\enspace \\{\rm where} \enspace q(\mathbf{d}^t_i|\mathbf{d}^0_i)&= {\rm Binomial}(k=\mathbf{d}^t_i,n=\mathbf{d}^0_i,p=\bar{\alpha}_t). \nonumber \\
    q(\mathbf{d}^t|\mathbf{d}^{t-1})&=\prod \limits_{i=1}^N q(\mathbf{d}^t_i|\mathbf{d}^{t-1}_i),\enspace \\ {\rm where} \enspace q(\mathbf{d}^t_i|\mathbf{d}^{t-1}_i)&= {\rm Binomial}(k=\mathbf{d}^t_i,n=\mathbf{d}^{t-1}_i,p=1-\beta_t). \nonumber
\end{align}

Intuitively, for $q(\mathbf{d}^t|\mathbf{d}^0)$, each of the $\mathbf{d}^0_i$ edges connected to node $i$ has a probability of $\bar{\alpha}$ of being retained at time step $t$. Therefore, the probability that the number of remaining edges equals $\mathbf{d}^t_i$ at time step $t$ follows a binomial distribution. A similar conclusion holds for $q(\mathbf{d}^t|\mathbf{d}^{t-1})$, except that the probability becomes $1-\beta_t$. 
Based on the above, we can now derive the posterior degree distribution in the reverse process, which is expressed as follows:
\begin{align}
    q(\mathbf{d}^{t-1}|\mathbf{d}^0,\mathbf{d}^t)&=\prod \limits_{i=1}^N q(\mathbf{d}^{t-1}_i|\mathbf{d}^t_i,\mathbf{d}^0_i),\enspace \\{\rm where} \enspace q(\mathbf{d}^{t-1}_i|\mathbf{d}^t_i,\mathbf{d}^0_i)&= {\rm Binomial}(k=\mathbf{d}^{t-1}_i-\mathbf{d}^t_i, \nonumber \\& \enspace \enspace \enspace n=\mathbf{d}^0_i-\mathbf{d}^t_i,p=\frac{\beta_t \bar{\alpha}_{t-1}}{1-\bar{\alpha}_t}). \nonumber
\end{align}

\begin{algorithm}[h]
    \caption{The Training Phase of GDDM}\label{algor:Training}

    \raggedright
    \textbf{Input}: The clean graph adjacency matrix $\textbf{A}^{0}$, diffusion schedule $\beta$, the learnable parameter $\theta$.\\
    % \textbf{Output}: The denoised graph adjacency matrix $\mathbf{A}^0$.

    \begin{algorithmic}[1] %[1] enables line numbers
        \WHILE{model convergence}
        \STATE Sample $t \thicksim \mathcal{U}[1,T]$, and obtain $t-1$.
        \STATE Draw $\mathbf{A}^{t-1} \thicksim q(\mathbf{A}^{t-1}|\mathbf{A}^{0})$ and $\mathbf{A}^{t} \thicksim q(\mathbf{A}^{t}|\mathbf{A}^{t-1})$.
        \STATE Obtain $\mathbf{s}^{t}$ with $\mathbf{A}^{t-1}$ and $\mathbf{A}^{t}$.
        \STATE Optimize the parameter $\theta$ by minimizing $\log\frac{p_{\theta}(\mathbf{A}^{t-1}|\mathbf{A}^t,\mathbf{s}^t)}{q(\mathbf{A}^{t-1}|\mathbf{A}^t,\mathbf{A}^0, \mathbf{s}^t)}$.
        \ENDWHILE
    \end{algorithmic}
\end{algorithm}

\begin{figure*}[!ht]
  \includegraphics[width=\textwidth]{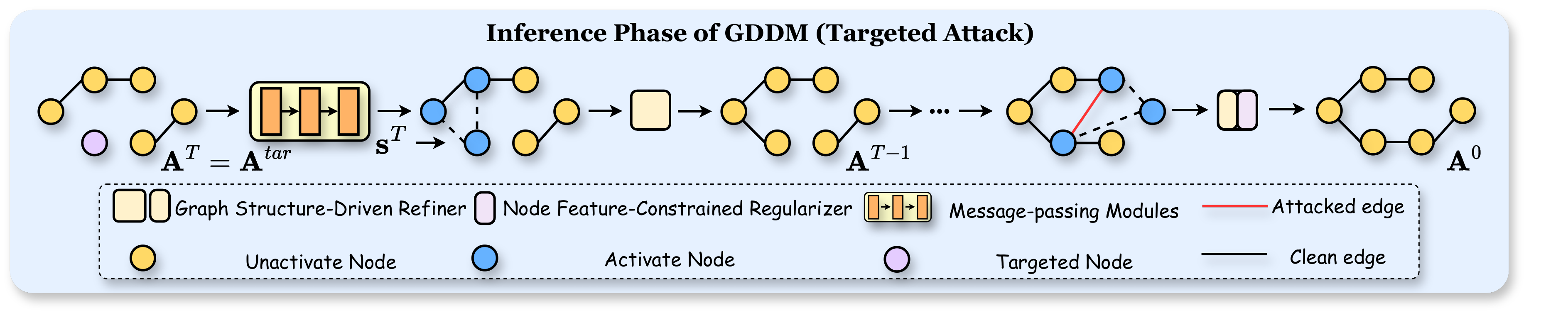}
  % \vskip -0.1in
  \Description{The Inference Phase}
  \caption{The Inference Phase of GDDM for Targeted Attacks. }
  % \vskip -0.1in 
  \label{fig:inference_process_for_targeted}
\end{figure*}

When a node's degree does not change between time steps $t-1$ and $t$, we always have $\mathbf{d}_i^t=\mathbf{d}_i^{t-1}$. When $\mathbf{d}_i^t=\mathbf{d}_i^{t-1}$,the probabilities of these events can be calculated by querying the degree distributions $q(\mathbf{d}^t|\mathbf{d}^{t-1})$ or $q(\mathbf{d}^{t-1}|\mathbf{d}^t,\mathbf{d}^0)$ as follows:
\begin{align}
    q(\mathbf{d}^t|\mathbf{d}^{t-1})&=(1-\beta_t)^{\mathbf{d}_i^{t-1}} \\
    q(\mathbf{d}^{t-1}|\mathbf{d}^0,\mathbf{d}^t)&=(1-\frac{\beta_t \bar{\alpha}_{t-1}}{1-\bar{\alpha}_t})^{\mathbf{d}^0_i-\mathbf{d}^t_i}
\end{align}

Let $\mathbf{s}^{t}_i$ be the random variable that node $i$ has degree change at time step $t$. Below we show the forward and reverse degree change distributions:

At timestep $t$, the forward degree change distribution for node $i$ given $\mathbf{d}^{t-1}_i$ is:
\begin{align}
    q(\mathbf{s}^{t}_i|\mathbf{d}^{t-1}_i)=\mathcal{B}(\mathbf{s}_i^t;1-(1-\beta_t)^{\mathbf{d}_i^{t-1}}).
\end{align}

At timestep $t$,  the reverse degree change distribution for node $i$ given $\mathbf{d}_i^0,\mathbf{d}_t^0$ is:
\begin{align}
q(\mathbf{s}^{t}_i|\mathbf{d}^{t}_i,\mathbf{d}^{0}_i)=\mathcal{B}(\mathbf{s}_i^t;1-(1-\frac{\beta_t \bar{\alpha}_{t-1}}{1-\bar{\alpha}_t})^{\mathbf{d}_i^{0}-\mathbf{d}_i^{t}}).
\end{align}

\begin{table}[h]
% \label{table3*}
\centering
% \vskip -0.1in
\renewcommand\arraystretch{1}
\caption{Statistics of the experimental data.}
% \vskip -0.1in
\label{tab:dataset}
\scalebox{1}{
\begin{tabular}{c|cccc}
    \toprule
    Dataset & N & E & Classes& Features\\
    \midrule
    Cora& 2,708 & 5,278 & 7& 1,433\\ 
    Citeseer& 3,327 & 4,552 & 6& 3,703\\  
    % Chameleon& 2,277 & 31,371 & 5&2,325\\
    Pubmed& 19,717 & 44,324 & 3&500\\
    \bottomrule
\end{tabular}}
% \vskip -0.2in
\end{table}

\subsection{Algorithm of Training Phase}\label{appendix:training_alg}
Graph diffusion models inherently involve high complexity in both time and space, posing significant challenges for efficient computation.
To address this challenge while preserving the model's ability to accurately reconstruct graph structures, we simplify the model by introducing a state vector $\mathbf{s}^{t}$, which tracks the nodes with degree changes at each timestep $t$ of the discrete graph diffusion process.
This simplification reduces computational overhead while maintaining the model's ability to restructure graphs effectively. 
The training phase of GDDM is shown in Algorithm~\ref{algor:Training}.

\subsection{Information Propagation Process Module} \label{appendix:detail_of_MPM}
In the information propagation process, the node representation $\mathbf{Z}$ and the representation $\mathbf{t}_{emb}$ of the timestep $t$ are concatenated as the input of GT or GNN. GRU adaptively retains important information inputting the output of GT and the nodes' hidden features $\mathbf{H}$:
\begin{align}
    \mathbf{Z}^{l} = {\rm concat}(\mathbf{Z}^l\|\mathbf{t}_{emb}\mathbf{1}^T),
\end{align}
\begin{align}
    \mathbf{Z}^{l} = {\rm GT}(\mathbf{Z}^{l}, \mathbf{A}^t) \hspace{0.5em}{\rm or} \hspace{0.5em} {\rm GNN}(\mathbf{Z}^{l}, \mathbf{A}^t),
\end{align}
\begin{align}
    \mathbf{Z}^l,\mathbf{H}^{l+1} = {\rm GRU}(\mathbf{Z}^{l},\mathbf{H}^{l}),
\end{align}
where $\mathbf{t}_{emb}$ is the representation of timestep $t$. $\mathbf{A}^t$  is the adjacency matrix of the graph at the current timestep $t$. 
${\rm GT}(\cdot)$ and ${\rm GNN}(\cdot)$ represent Graph Transformer and Graph Neural Network, which perform attentive information propagation between nodes, allowing each node to aggregate local feature information from its neighborhood in the graph. ${\rm GRU}(\cdot)$ represents Gated Recurrent Unit (GRU).

% \noindent\textbf{Selective information retention.}
% To adaptively retain important information during the message-passing process, the updated node representation and the node's hidden state features $\mathbf{H}^{l}$ in the corresponding $l$-th MPM are fed into the GRU, resulting in the hidden state features for the next MPM and the node representation:

% where ${\rm GRU}(\cdot)$ is a Gated Recurrent Unit (GRU).

% \noindent\textbf{Global context update.}
To update the global contextual feature $\mathbf{c}^l$, the representation of each node and the current global contextual feature $\mathbf{c}^l$ are concatenated and applied a mean operation as the input of a MLP. The output of the MLP serves as the global contextual feature for the next layer:
\begin{align}
    \mathbf{c}^{l+1} = {\rm MLP}({\rm mean}({\rm concat}(\mathbf{Z}^l\|\mathbf{c}^l\mathbf{1}^T))).
\end{align}

% \noindent\textbf{Node representation update.}
Finally, the node representations $Z$ and the global contextual feature $\mathbf{c}^{l+1}$ are summed to obtain the initial node representations for the next MPM:
\begin{align}
    \mathbf{Z}^{l+1} = \mathbf{Z}^{l} + \mathbf{c}^{l+1}\mathbf{1}^T.
\end{align}

\subsection{The Inference Phase of GDDM for Targeted Attack}\label{appendix:denoise_for_targeted}
GDDM adopts tailored strategies for different types of adversarial attacks. 
For \textbf{targeted attacks}, where specific nodes are compromised, a more focused strategy is necessary. GDDM starts by removing all edges connected to the target nodes, constructing an initial adjacency matrix $\mathbf{A}^{tar}$.
The detail of the Tailored Attack-specific Denoising Strategies (TADS) is shown in Figure~\ref{fig:inference_process_for_targeted}.

% \subsection{Algorithm of The GDDM Inference Phase} \label{appendix:alg_inference}
% Following the completion of the training phase, GDDM applies the learned diffusion dynamics to reconstruct the clean graph from the attacked one.
% The overall inference phase of GDDM is shown in Algorithm~\ref{algor:Inference}.

\section{Experiments Detail}
\subsection{Dataset}\label{appendix:dataset}
The statistics are listed in Tabel~\ref{tab:dataset}.
Cora, Citeseer and Pubmed are citation networks where nodes represent papers from seven,six and three research categories, respectively, and edges denote citation links. 
% Chameleon is a Wikipedia graph with edges representing mutual links and nodes often connect to different classes.

\subsection{Baselines}\label{appendix:baseline}
\begin{itemize}
    \item \textbf{GCN}: GCN is a widely used graph convolutional network grounded in spectral graph theory.
    It applies convolution operations on graphs to capture the relationships between nodes based on their connectivity.
    \item \textbf{Jaccard}: Jaccard prunes edges based on the feature similarity between nodes, as adversarial attacks tend to connect dissimilar nodes in the graph.
    \item \textbf{SVD}: SVD removes the high-rank perturbations introduced by adversarial attacks by the low-rank approximation of the attacked graph, enhancing the robustness of GNN model.
    \item \textbf{ProGNN}: ProGNN mitigates the negative impact of adversarial attacks on GNN models by optimizing the adjacency matrix of the attacked graph using regularization terms based on feature smoothness, low-rank, and sparsity.
    \item \textbf{Guard}: Guard uses a scoring function based on node degree and gradient to identify anchor nodes (attacker nodes), then removes all edges between the anchor nodes and the target node to ensure the target node is protected from attacks.
    % \item \textbf{Rand}: Rand is a baseline version of Guard, where the strategy for selecting anchor nodes is replaced by random selection instead of using a scoring function.
    % \item \textbf{Deg}: Deg is another baseline version of Guard, where the strategy for selecting anchor nodes is based on node degree, as lower-degree nodes are more likely to become anchor nodes.
    \item \textbf{STABLE}: STABLE is an unsupervised method for optimizing graph structure, which leverages a specially designed robust GCN to effectively defend against adversarial attacks on graph data.
    \item \textbf{SG-GRS}: SG-GSR is a self-guided framework that enhances robustness against adversarial attacks by extracting a clean subgraph from the attacked graph itself.
    \item \textbf{WGTL}: WGTL improves GNN robustness by integrating local and global topological features through witness complexes and persistent homology. In this paper, we use GCN combined with WGTL as a baseline.
    \item \textbf{DiffSP}: DiffSP is a defense method that models both attack and purification through a unified graph diffusion process, enabling robust classification under structural perturbations.
    \item \textbf{GraD}: GraD introduces an improved attack objective function that enables the attack model to target medium-confidence nodes, thereby increasing attack efficiency.
    \item \textbf{Nettack}: Nettack is a greedy-based attack method that perturbs targeted nodes in a graph while preserving the node degree distribution and feature co-occurrence.
    
\end{itemize}

\subsection{Implementation Details}\label{appendix:implementation_details}
We generate attacks on each graph based on the perturbation number$/$rate.
% and we ensure that all hyper-parameters in the attack methods align with the original authors’ implementations.
During the model training phase, we primarily adjust the diffusion steps, with a range of $\{64, 128, 256, 512\}$.
During the model inference phase, we primarily adjust the expected size of the generated graph, ranging from 70\% to 95\%.
Performance is reported as the average accuracy with standard deviation, based on 10 runs on clean, perturbed or denoised graphs.
We closely follow the default setting of the baselines for a fair comparison.

\end{document}